\documentclass[twoside,11pt]{article}

\usepackage{blindtext}

\usepackage{caption}
\usepackage{subcaption}

\usepackage[preprint]{jmlr2e}

\usepackage[utf8]{inputenc}
\usepackage{color}
\usepackage{float}
\usepackage{bm}  
\usepackage{hyperref}       
\usepackage{url}            
\usepackage{booktabs}       
\usepackage{tabularx}
\usepackage{amsfonts}       
\usepackage{amsmath}
\usepackage{nicefrac}       
\usepackage{microtype}      
\usepackage{tikz}           
\usepackage{centernot}
\usepackage[ruled,vlined,linesnumbered]{algorithm2e}
\usepackage{mathtools}
\usepackage{enumitem}

\graphicspath{{figs/}}

\jmlrheading{00}{0000}{1-\pageref{LastPage}}{6/25; Revised 0/00}{0/00}{00-0000}{Ujas Shah, Manuel Lladser and Rebecca Morrison}
\usepackage{lastpage}

\ShortHeadings{Conditional Independence Estimates for the Generalized Nonparanormal}{Shah, Lladser and Morrison}
\firstpageno{1}

\begin{document}

\title{Conditional Independence Estimates \\ for the Generalized Nonparanormal}

\author{\name Ujas Shah \email ujas.shah@colorado.edu \\
       \addr Department of Computer Science\\
       University of Colorado\\
       Boulder, CO 80309-0430, USA
       \AND
       \name Manuel Lladser \email manuel.lladser@colorado.edu \\
       \addr Department of Applied Mathematics\\
       University of Colorado\\
       Boulder, CO 80309-0526, USA
       \AND
       \name Rebecca Morrison \email rebeccam@colorado.edu \\
       \addr Department of Computer Science\\
       University of Colorado\\
       Boulder, CO 80309-0430, USA}

\editor{My editor}

\maketitle

\begin{abstract}
For general non-Gaussian distributions, the covariance and precision matrices do not encode the independence structure of the variables, as they do for the multivariate Gaussian. This paper builds on previous work to show that for a class of non-Gaussian distributions---those derived from diagonal transformations of a Gaussian---information about the conditional independence structure can still be inferred from the precision matrix, provided the data meet certain criteria, analogous to the Gaussian case. We call such transformations of the Gaussian as the \textit{generalized nonparanormal}. The functions that define these transformations are, in a broad sense, arbitrary. We also provide a simple and computationally efficient algorithm that leverages this theory to recover conditional independence structure from the generalized nonparanormal data. The effectiveness of the proposed algorithm is demonstrated via synthetic experiments and applications to real-world data.
\end{abstract}


\section{Introduction}

There are not many common parametric multivariate continuous distributions. Of course, the multivariate normal reigns supreme, and for good reasons: arguments based on the central limit theorem and the principle of maximum entropy often result in the Gaussian distribution. Others include the uniform hypercube, multivariate variations of the Student-$t$, and Gaussian mixtures; but these are all applicable in rather specific settings. Empirical distributions often do not fit in any of these groups, and it is of general interest to mathematically (parametrically) describe general non-Gaussian multivariate distributions.

In this paper, we are concerned with the problem of inferring the independence structure of non-Gaussian distributions from data. A common way to do this is by using a Gaussian copula~\citep{nelsen2007introduction} or assuming that the data follows a nonparanormal distribution~\citep{liu2009nonparanormal}, which are distributions obtained by applying marginal \emph{monotone} transformations to a Gaussian random vector.

In the nonparanormal case, take the non-Gaussian vector $\bm Z = (Z_1, \dots, Z_d)$ and apply monotone marginal transformations $g_i$, so that $\bm X = (g_1(Z_1),\dots,g_d(Z_d))\sim \mathcal{N}(\bm 0, \bm \Sigma_\rho)$. In what follows, $\rho$ denotes the distribution of this centered multivariate normal. (The mean of this distribution can be set arbitrarily, however, we assume without any loss of generality that it is $\bm 0$.) In this way, the relationship between the original variables is captured by the (Gaussian) covariance matrix, $\bm \Sigma_\rho$, while each $g_i$ can define an arbitrary marginal for $X_i$. This description conveniently captures marginal and conditional independence properties---a zero in the $ij$-th entry of $\bm \Sigma_\rho$ means that variables $i$ and $j$ are marginally independent, while a zero in the $ij$-th entry of $\bm \Sigma_\rho^{-1}$ means that the two are conditionally independent, given a separator set. In short: \textit{$\bm X$ and $\bm Z$ obey the same independence structure.}

If this structure is unknown, one must determine a graphical model from expert knowledge or directly from the data. To give just a few application examples, \citet{dong2015laplacian}~builds graphical models for signal processing, by learning graph Laplacians that capture the underlying topology of the data. In neuroscience, \citet{smith2011network}~applies graphical models to fMRI data to estimate brain connectivity networks. In finance, \citet{mastromatteo2012reconstruction}~constructs financial networks to estimate systemic risk, using graphical models to represent dependencies between financial institutions. As a final example, \citet{yang2011like}~introduces methods for learning the structure of social networks by modeling people's interests, leveraging graphical models for community detection.

In learning scenarios for the nonparanormal distribution, the parameters $\bm \Sigma_\rho$ and the transformation $\bm g=(g_1,\ldots,g_d)$ are estimated from data, thereby giving a complete semi-parametric specification of $\bm Z$~\citep{liu2009nonparanormal}. Note that doing so relies essentially on density estimation since the set of parameters $(\bm \Sigma_\rho, \bm g)$ fully specifies $\bm Z$. Other methods have been developed that do not require estimation of $\bm g$. A rank-based algorithm estimates the covariance matrix using monotone-transformation invariant measures like Spearman's $\rho$ and Kendall's $\tau$, which is then inverted to get the precision matrix, $\bm \Gamma = \bm \Sigma_\rho^{-1}$ ~\citep{liu2012nonparanormal, xue2012regularized}. Rank correlation measures have also been used to modify existing algorithms such as the Peter-Clark algorithm ~\citep{spirtes2001causation} and the Grow-Shrink algorithm ~\citep{margaritis2003learning} to learn the independence structure of nonparanormal data ~\citep{harris2013pc, musella2019copula}.

For more general distributions, methods to recover the independence structure for exponential-type tailed and polynomial-type tailed distributions through linear programming have been proposed ~\citep{cai2011constrained}. Recent work has explored node regression combined with projection pursuit to construct graphical models for generalized multivariate skew-normal distributions~\citep{nghiem2022estimation}. Another approach (by two of the co-authors) identifies conditional independence for arbitrary non-Gaussian distributions from the Hessian of the log density, but is (so far) computationally limited to rather small graphs~\citep{baptista2024learning, morrison2017beyond}.

In this paper, we show that the marginal and conditional independence properties of a broad set of distributions can be estimated directly, \emph{bypassing density estimation}, provided the distributions meet certain criteria. We consider functions $f_i$, and inspect the distribution of the random vector $\bm Z = (f_1(X_1),\dots, f_d(X_d))$, where $\bm X\sim\rho$. (In the context of the previous notation, $g_i = f_i^{-1}$ when $f_i$ is invertible, although this constraint is not required in our analysis.) We say that $\bm Z$ is a \emph{generalized nonparanormal} (GNPN) distribution and denote this distribution as $\pi$; namely, $\bm Z\sim\pi$. This setup is illustrated with an example in Section~\ref{ssec:ex}.

This work is an extension on a previous paper by one of the co-authors~\citep{morrison2022diagonal}. That paper assumed that the functions $f_i$ were odd, that each is the same, i.e., $f_i = f$ for all $i$, and that $f$ is smooth and has uniformly bounded derivatives at 0. More recently, these assumptions have been relaxed further allowing the functions to be even, odd, or neither, and each can be different and possibly discontinuous~\citep{morrison2024exact}. The GNPN distribution can be viewed, as the name suggests, as a generalization of the nonparanormal, where the $f_i$ are not restricted to being monotonic functions.

\subsection{Example}\label{ssec:ex} 
Consider a circle graph with eight nodes and precision matrix
$$\bm \Gamma_\rho = \left(
\begin{array}{cccccccc}
 1 & \alpha & 0 & 0 & 0 & 0 & 0 & \alpha \\
 \alpha & 1 & \alpha & 0 & 0 & 0 & 0 & 0 \\
 0 & \alpha & 1 & \alpha & 0 & 0 & 0 & 0 \\
 0 & 0 & \alpha & 1 & \alpha & 0 & 0 & 0 \\
 0 & 0 & 0 & \alpha & 1 & \alpha & 0 & 0 \\
 0 & 0 & 0 & 0 & \alpha & 1 & \alpha & 0 \\
 0 & 0 & 0 & 0 & 0 & \alpha & 1 & \alpha \\
 \alpha & 0 & 0 & 0 & 0 & 0 & \alpha & 1 \\
\end{array}
\right).$$
Let $\alpha = 1/22$. Then its associated covariance matrix, i.e., inverse (to 4 decimal places) is found to be
$$\bm \Sigma_\rho = \left(
\begin{array}{rrrrrrrr}
 1.0042 & -0.0457 & 0.0021 & -0.0001 & 0.0\,\,\,\,\, & -0.0001 & 0.0021 & -0.0457 \\
 -0.0457 & 1.0042 & -0.0457 & 0.0021 & -0.0001 & 0.0\,\,\,\,\, & -0.0001 & 0.0021 \\
 0.0021 & -0.0457 & 1.0042 & -0.0457 & 0.0021 & -0.0001 & 0.0\,\,\,\,\, & -0.0001 \\
 -0.0001 & 0.0021 & -0.0457 & 1.0042 & -0.0457 & 0.0021 & -0.0001 & 0.0\,\,\,\,\, \\
 0.0\,\,\,\,\, & -0.0001 & 0.0021 & -0.0457 & 1.0042 & -0.0457 & 0.0021 & -0.0001 \\
 -0.0001 & 0.0\,\,\,\,\, & -0.0001 & 0.0021 & -0.0457 & 1.0042 & -0.0457 & 0.0021 \\
 0.0021 & -0.0001 & 0.0\,\,\,\,\, & -0.0001 & 0.0021 & -0.0457 & 1.0042 & -0.0457 \\
 -0.0457 & 0.0021 & -0.0001 & 0.0\,\,\,\,\, & -0.0001 & 0.0021 & -0.0457 & 1.0042 \\
\end{array}\right).$$

We transform centered Gaussian data created with the above covariance matrix; for simplicity, let $f_i(x) = x^3 \,\,\forall i$. The resulting data have the following sample covariance matrix, computed from 100,000 samples:
$$\bm{\hat \Sigma_\pi} = \left(
\begin{array}{rrrrrrrr}
15.3436 & -0.4298 & -0.0375 & 0.0065 & 0.0447 & 0.0172 & 0.0701 & -0.3848 \\ 
-0.4298 & 15.1577 & -0.3992 & 0.0467 & -0.032 & -0.0693 & 0.0793 & -0.019 \\ 
-0.0375 & -0.3992 & 15.1956 & -0.4166 & 0.0131 & 0.0015 & 0.0561 & 0.0056 \\ 
0.0065 & 0.0467 & -0.4166 & 15.3007 & -0.3941 & 0.0032 & 0.0778 & -0.0375 \\ 
0.0447 & -0.032 & 0.0131 & -0.3941 & 15.4689 & -0.4228 & 0.0193 & 0.0167 \\ 
0.0172 & -0.0693 & 0.0015 & 0.0032 & -0.4228 & 15.5416 & -0.3632 & 0.033\,\,\, \\
0.0701 & 0.0793 & 0.0561 & 0.0778 & 0.0193 & -0.3632 & 15.6811 & -0.4076 \\ 
-0.3848 & -0.019 & 0.0056 & -0.0375 & 0.0167 & 0.033 & -0.4076 & 15.2842 \\ 
\end{array}
\right).$$

We also computed the exact covariance matrix as described from the theory in sections~\ref{sec:cov}~and~\ref{sec:est_gamma}, which yielded the following:
$$\bm \Sigma_\pi = \left(
\begin{array}{rrrrrrrr}
15\,\,\,\,\,\, & -0.4156 & 0.0189 & -0.0009 & 0.0001 & -0.0009 & 0.0189 & -0.4156 \\
-0.4156 & 15\,\,\,\,\,\, & -0.4156 & 0.0189 & -0.0009 & 0.0001 & -0.0009 & 0.0189 \\
0.0189 & -0.4156 & 15\,\,\,\,\,\, & -0.4156 & 0.0189 & -0.0009 & 0.0001 & -0.0009 \\
-0.0009 & 0.0189 & -0.4156 & 15\,\,\,\,\,\, & -0.4156 & 0.0189 & -0.0009 & 0.0001 \\
0.0001 & -0.0009 & 0.0189 & -0.4156 & 15\,\,\,\,\,\, & -0.4156 & 0.0189 & -0.0009 \\
-0.0009 & 0.0001 & -0.0009 & 0.0189 & -0.4156 & 15\,\,\,\,\,\, & -0.4156 & 0.0189 \\
0.0189 & -0.0009 & 0.0001 & -0.0009 & 0.0189 & -0.4156 & 15\,\,\,\,\,\, & -0.4156 \\
-0.4156 & 0.0189 & -0.0009 & 0.0001 & -0.0009 & 0.0189 & -0.4156 & 15\,\,\,\,\,\, \\
\end{array}
\right).$$
(Note that the matrix above is quite similar to $\bm{\hat \Sigma_\pi}$.)

Finally, the precision matrix, computed as the inverse of the theoretical covariance, is: 
$$\bm \Gamma_\pi = \left(
\begin{array}{rrrrrrrr}
0.0668 & 0.0018 & -0.0\,\,\,\,\, & 0.0\,\,\,\,\, & -0.0\,\,\,\,\, & 0.0\,\,\,\,\, & -0.0\,\,\,\,\, & 0.0018 \\
0.0018 & 0.0668 & 0.0018 & -0.0\,\,\,\,\, & 0.0\,\,\,\,\, & -0.0\,\,\,\,\, & 0.0\,\,\,\,\, & -0.0\,\,\,\,\, \\ 
-0.0\,\,\,\,\, & 0.0018 & 0.0668 & 0.0018 & -0.0\,\,\,\,\, & 0.0\,\,\,\,\, & -0.0\,\,\,\,\, & 0.0\,\,\,\,\, \\ 
0.0\,\,\,\,\, & -0.0\,\,\,\,\, & 0.0018 & 0.0668 & 0.0018 & -0.0\,\,\,\,\, & 0.0\,\,\,\,\, & -0.0\,\,\,\,\, \\ 
-0.0\,\,\,\,\, & 0.0\,\,\,\,\, & -0.0\,\,\,\,\, & 0.0018 & 0.0668 & 0.0018 & -0.0\,\,\,\,\, & 0.0\,\,\,\,\, \\ 
0.0\,\,\,\,\, & -0.0\,\,\,\,\, & 0.0\,\,\,\,\, & -0.0\,\,\,\,\, & 0.0018 & 0.0668 & 0.0018 & -0.0\,\,\,\,\, \\ 
-0.0\,\,\,\,\, & 0.0\,\,\,\,\, & -0.0\,\,\,\,\, & 0.0\,\,\,\,\, & -0.0\,\,\,\,\, & 0.0018 & 0.0668 & 0.0018 \\ 
0.0018 & -0.0\,\,\,\,\, & 0.0\,\,\,\,\, & -0.0\,\,\,\,\, & 0.0\,\,\,\,\, & -0.0\,\,\,\,\, & 0.0018 & 0.0668 \\
\end{array}
\right).$$

The diagonal entries are now $0.0668$, the off-diagonal edge entries $0.0018$, and all other entries (corresponding to no edge) only have a non-zero value in the fifth decimal place onward, at most. Importantly for learning the dependence structure of GNPN data, there are two orders of magnitude separating edges and non-edges for this particular transformation. This is the behavior we will exploit to learn the conditional dependence structure.

\begin{figure}[tbp]
    \center
    \includegraphics[width=.9\textwidth]{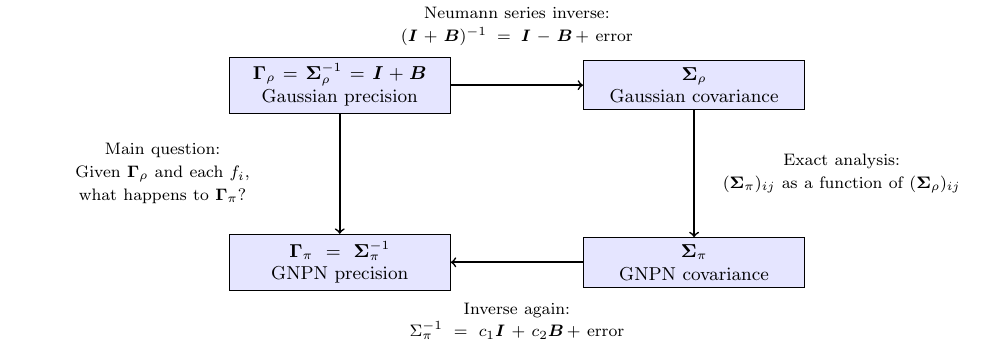} 
    \caption{We use $\rho$ and $\pi$ to represent the Gaussian and generalized nonparanormal (GNPN) distributions, respectively. $\bm \Gamma = \bm \Sigma^{-1}$ is the inverse covariance or the precision matrix of the corresponding distribution. The analysis along the right downward arrow, from the Gaussian covariance ($\bm \Sigma_\rho$) to the GNPN covariance ($\bm \Sigma_\pi$), often results in exact answers; that theory is given in~\protect\citet{morrison2024exact}. In this paper, we track what happens along both the horizontal arrows, along with the right downward arrow, to answer the main question (left downward arrow): what can one say about the \(\bm \Gamma_\pi\) directly from \(\bm \Gamma_\rho\) and each \(f_i\). 
    \label{fig:flow}}
\end{figure}

The remainder of the paper is organized as follows. In Section~\ref{sec:cov}, we estimate the entries of $\bm \Sigma_\pi$, the covariance of the GNPN distribution, given the pre-transformed Gaussian. In Section~\ref{sec:est_gamma}, we provide rigorous estimates for the entries of $\bm \Gamma_\pi$, the precision of the GNPN distribution. Based on this theory, we show that, under some assumptions, the Markov properties of the GNPN distribution are discernible from its precision and covariance matrix. In other words, the Markov properties can be learned through matrix estimation, analogous to the Gaussian case. The analysis so far can be summarized as Figure~\ref{fig:flow}. Section~\ref{sec:algo} describes a simple algorithm to identify the conditional independence graph for some GNPN data. Section~\ref{sec:ex} details an example of the algorithm applied to recover a graph and also tests the algorithm's performance with simulations and real data. Section~\ref{sec:prac_considerations} discussed a check to identify whether the algorithm, and the analysis, are applicable to some given data as well as the algorithm's sample efficiency. Finally, we conclude with Section~\ref{sec:dis}
 
\section{Moments after General Diagonal Transformations}\label{sec:cov} 

In this section, we restate some necessary results from \cite{morrison2024exact}. The specifics vary, depending on whether the functions are restricted to be odd, same in each variable, and/or continuous. We focus here on the case that the functions $f_i$, $i=1,\ldots,d$, satisfy a derivative constraint and can be different in each variable. In particular, there is no constraint on the parity of the functions. The other cases can be derived similarly. 

Let $f_i$ be a smooth function of $x$ with derivatives at 0 satisfying the condition
\begin{equation}\label{der.est} 
|f_i^{(a)}(0)| \leq C_i\,K_i^a, 
\end{equation} 
for some (finite) constants $C_i$ and $K_i$ and all integer $a\ge0$. (Notice that if $|K_i| \leq 1$, the derivatives of $f_i$ are uniformly bounded.) In particular, for each integer $k\ge0$, the following Taylor series has an infinite radius of convergence:
\[F_{ki}(x) := \sum_{u\geq 0} \frac{f_i^{(2u + k)}(0)}{u!} x^u = f_i^{(k)}(0)+f_i^{(k+2)}(0) x+\frac{f_i^{(k+4)}(0)}{2!} x^2+\ldots\]
In particular, $F_{ki}$ is infinitely differentiable, so all of its derivatives are continuous.

In all that follows, let $\sigma_{ij} := (\Sigma_\rho)_{ij}$ and $\tau_{ij} := (\Sigma_\pi)_{ij}$. Furthermore, let
\[G_{kij}(x) := F_{ki}(\sigma_{ii}x)\,F_{kj}(\sigma_{jj}x).\]

The following theorem is adapted from \cite{morrison2024exact} to allow for $f_i \neq f_j$.

\begin{theorem}{\label{main.th}}
If each $f_i$ satisfies condition~(\ref{der.est}) then

\[\tau_{ij} = \sum_{k \geq 1} G_{kij}(1/2)\frac{\sigma_{ij}^k}{k!}
= G_{1ij}(1/2) \sigma_{ij} + \frac{1}{2}\,G_{2ij}(1/2) \sigma_{ij}^2 +\mathcal{O}(\sigma_{ij}^3),\]
where the hidden constant for the $O$ estimate depends only on $f_i, f_j, \sigma_{ii},$ and $ \sigma_{jj}$.
\end{theorem}

\section{Estimates of the Generalized Nonparanormal Precision, $\Gamma_\pi$}\label{sec:est_gamma}

From now on, $\|x\|:=\sqrt{x{'}x}$ with $x\in\mathbb{R}^d$ denotes the Euclidean norm of a vector $x\in\mathbb{R}^d$. Similarly, $\|A\|$ denotes the operator norm of a $d\times d$ real matrix $A=(A_{ij})$ with respect to the Euclidean norm. Namely:
\[\|A\|:=\sup_{x\in\mathbb{R}^d, x \neq 0}\frac{\|Ax\|}{\|x\|}.\]
Instead, $\|\bm A\|_{\text{HS}}$ denotes the Hilbert-Schmidt norm of $A$, i.e.:
\[\|\bm A\|_{\text{HS}}=\sqrt{\sum_{i,j=1}^d A_{ij}^2}.\]
It is well-known that $\|A\|\le\|A\|_{\text{HS}}$.

In this section, we build a series of results to reach an estimate for $\bm \Gamma_\pi$ in terms of $\bm \Gamma_\rho$. The first lemma,~\ref{tau:est}, estimates $\tau_{ij}$ for $i \neq j$ in terms of $\sigma_{ij}$ and a multiplying factor $G_{1ij}(1/2) = F_{1i}(\sigma_{ii}/2)F_{1j}(\sigma_{jj}/2)$; so, to first order, the approximation still depends on $\sigma_{ii}$ and $\sigma_{jj}$ only. Lemma~\ref{lem:kap} estimates $\tau_{ij}$ in terms of just $\sigma_{ij}$ but restricted to the case that $i = j$. Finally, Lemma~\ref{lem:tau2} estimates $\tau_{ij}$ in terms of $\sigma_{ij}$ for $i \neq j$---the estimate is not as fine as that of the first lemma, but the result no longer depends on any other entries of the original covariance besides $\sigma_{ij}$. Thus we can map elements from $\bm \Sigma_\rho$ to $\bm \Sigma_\pi$ entry-wise.

\begin{lemma}\label{tau:est} Let $i \neq j$. Suppose that $f_i$, $f_j$ satisfy condition~(\ref{der.est}) and that $\Gamma_\rho = I + B$, with $\|B\| \leq \delta <1/2$. If $\epsilon := \frac{\delta}{1-\delta}$ then  \[\tau_{ij} = G_{1ij}(1/2) \sigma_{ij} + \mathcal{O}(\epsilon^2).\]
\end{lemma}
\proof
First, since \(\Sigma_\rho = \Gamma_\rho^{-1} = (I+B)^{-1}\), the Neumann series of \((I+B)^{-1} = I-B^{'}\), where \(B^{'} = B - B^2 + B^3 - \dots\). Because \(\|B\|\le\delta\), we have
\[\|B^{'}\| \le \|B\| + \|B^2\| + \|B^3\| + \dots \le \delta + \delta^2 + \dots = \frac{\delta}{1-\delta} = \epsilon .\]
Thus, due to the Cauchy–Schwarz inequality, each off-diagonal entry of $\Sigma_\rho$ satisfies $|\sigma_{ij}| = |B^{'}_{ij}| \leq \epsilon < 1$.

Now, let $C_i$, $K_i$ and $C_j$, $K_j$ be the constants associated with $f_i$ and $f_j$ in condition~(\ref{der.est}), respectively. Let $C := \max(C_i, C_j)$ and $K := \max(K_i, K_j)$. Then, due to Theorem \ref{main.th}, we find that
\begin{align*}
    \left| \tau_{ij} - G_{1ij}(1/2) \sigma_{ij} \right| &\leq \sum_{k \geq 2} \left| G_{kij}(1/2)
\frac{\sigma_{ij}^k}{k!}\right|\\
    &\leq C^2 \epsilon^2 e^{K^2(\sigma_{ii} +\sigma_{jj})/2} \sum_{k \geq 2} \frac{K^{2k}\left| \sigma_{ij}\right|^{k-2}}{k!}\\
    &\leq C^2 \epsilon^2 \,e^{K^2(\sigma_{ii} +\sigma_{jj} +2)/2}.
\end{align*} \hfill 
\endproof

\begin{lemma}\label{lem:kap}
    Suppose $f_i$ satisfies condition~(\ref{der.est}) and that $\Gamma_\rho = I + B$, with $\|B \| \leq \delta < 1/2$ and $B_{ii} = 0$. If $\kappa_i := \sum_{k\geq 1} \frac{F_{ki}^{2}(1/2)}{k!}$ and $\epsilon := \frac{\delta}{1-\delta}$ then
    \[\tau_{ii} = \kappa_i \sigma_{ii} + \mathcal{O}(\epsilon^2).\]
\end{lemma}
\proof
Due to Theorem~\ref{main.th}:
\[\tau_{ii} 
= \sum_{k\geq 1}G_{kii}(1/2) \frac{\sigma_{ii}^k}{k!}
= \sum_{k\geq 1}F_{ki}^2(1/2) \frac{\sigma_{ii}^k}{k!}.\]
The above identity lets us think of $\tau_{ii}$ as a function of $\sigma_{ii}$; in particular, $\kappa_i=\tau_{ii}(1)$. Moreover, $\tau_{ii}$ and any of its derivatives are continuous functions of $\sigma_{ii}$ because the above series has an infinite radius of convergence due to condition~(\ref{der.est}).

On the other hand, the Neumann series of $(I+B)^{-1}$ implies that $\Sigma_\rho=I-B+\mathcal{O}(\delta\epsilon)$; in particular, $\sigma_{ii} = 1 + \mathcal{O}(\epsilon^2)$ because $B_{ii} = 0$ and $0\le\delta\le\epsilon$. Hence, expanding $\tau_{ii}$ around $\sigma_{ii}=1$ we obtain that
$\left|\tau_{ii}(\sigma_{ii}) - \tau_{ii}(1) \right| \leq \max_{1 \in (a,b)} \left|
\tau_{ii}'(x)\right| \left| \sigma_{ii} - 1 \right|$, where $(a,b)$ is any interval containing $1$ and $\sigma_{ii}$. So
\begin{align*}\tau_{ii} 
&= \kappa_i + \mathcal{O}(\epsilon^2)\\
    &= \kappa_i\sigma_{ii} + \kappa_i(1 - \sigma_{ii}) + \mathcal{O}(\epsilon^2)\\
    &= \kappa_i\sigma_{ii} + \mathcal{O}(\epsilon^2).
\end{align*} \hfill 
\endproof

\begin{lemma}\label{lem:tau2}
Let $i \neq j$. Suppose that $f_i$, $f_j$ satisfy condition~(\ref{der.est}) and that $\Gamma_\rho = I + B$, with $\|B \| \leq \delta < 1/2$ and $B_{ii}=B_{jj}=0$. If $\lambda_{ij} := F_{1i}(1/2)F_{1j}(1/2)$ and $\epsilon := \frac{\delta}{1-\delta}$ then 
\[\tau_{ij} = \lambda_{ij} \sigma_{ij} + \mathcal{O}(\epsilon^2).\] 
\end{lemma}
\proof
This proof is similar to the previous one, but here we expand the function $F_{1i}(x/2)$ around $x=\sigma_{ii}$ to obtain that
$\left| F_{1i}(\sigma_{ii}/2) - F_{1i}(1/2) \right| \leq \alpha \left| \sigma_{ii} - 1 \right|$ 
for some finite constant $\alpha$. As \(\sigma_{ii} = 1 + \mathcal{O}(\epsilon^2)\) and \(\sigma_{jj} = 1 + \mathcal{O}(\epsilon^2)\) because $B_{ii}=0$ and $B_{jj}=0$, respectively, we have 
$F_{1i}(\sigma_{ii}/2)=F_{1i}(1/2)+\mathcal{O}(\epsilon^2)$ and $F_{1j}(\sigma_{jj}/2)=F_{1j}(1/2)+\mathcal{O}(\epsilon^2)$. Hence
\begin{align*}
    G_{1ij}(1/2) &= F_{1i}(\sigma_{ii}/2)F_{1j}(\sigma_{jj}/2)\\
    &=\left(F_{1i}(1/2) + \mathcal{O}(\epsilon^2) \right) \left( F_{1j}(1/2) +
    \mathcal{O}(\epsilon^2) \right)\\
    &= F_{1i}(1/2)F_{1j}(1/2) + \mathcal{O}(\epsilon^2),
\end{align*}
and Lemma~\ref{tau:est} finally implies that $\tau_{ij} = F_{1i}(1/2)F_{1j}(1/2)\sigma_{ij} + \mathcal{O}(\epsilon^2)$.
 \endproof

Due to the previous results, we can assert the following structure for the transformed covariance $\bm{\Sigma}_\pi$.

\begin{theorem}
\label{thm:main}
    Let $d$ be a fixed dimension. If $\bm \Gamma_\rho = \bm I + \bm B$, with $\| \bm B\| \leq \delta < 1/2$, and $B_{ii} = 0$ for each $i$, then 
    \[\bm \Sigma_\pi = \bm{\mathrm{K}} \bm I - \bm \Lambda \bm B \bm \Lambda + \bm E^{'},\]
    where $\bm{\mathrm{K}} := \text{diag}(\kappa_1, \dots, \kappa_d)$, $\bm \Lambda :=
\text{diag}(\lambda_1, \dots, \lambda_d)$ with $\lambda_i := F_{1i}(1/2)$, and $\|\bm E^{'}\|=\mathcal{O}(\epsilon^2)$.
\end{theorem} \hfill
\proof
All of the entries of $\bm E^{'}$ are at most $\mathcal{O}(\epsilon^2)$. Thus, for fixed $d$, $\|\bm E^{'} \|  \leq \|\bm E^{'} \|_{\text{HS}} = \mathcal{O}(d \epsilon^2) =\mathcal{O}(\epsilon^2)$.
\endproof

\begin{remark}
A tighter error-bound is possible in the theorem if each function $f_i$ is odd, in which case $\|\bm E^{'} \| \leq \|\bm E^{'} \|_{HS}=\mathcal{O}(d \epsilon^3)$. In particular, $\|\bm E^{'} \|=\mathcal{O}(\epsilon^2)$ when $d \epsilon$ stays bounded.
\end{remark}

Our last step in this section is to estimate the entries of the transformed precision matrix. For this, under the hypothesis of Theorem~\ref{thm:main}, we have that
\begin{align*}
    \bm \Gamma_\pi 
    &= \bm  \Sigma_\pi^{-1} \\
        &= \left(\bm I +\left(- \bm{\mathrm{K}}^{-1} \bm \Lambda \bm B \bm \Lambda + \bm{\mathrm{K}}^{-1} \bm E^{'}\right) \right)^{-1}\bm{\mathrm{K}}^{-1}\\
        &= \left(\bm I + \bm{\mathrm{K}}^{-1} \bm \Lambda \bm B \bm \Lambda - \bm{\mathrm{K}}^{-1} \bm E^{'} + \bm E^{''}\right)\bm{\mathrm{K}}^{-1}\\
        &= \bm{\mathrm{K}}^{-1} \bm I + \bm{\mathrm{K}}^{-1} \bm \Lambda \bm B \bm \Lambda \bm{\mathrm{K}}^{-1} 
        + \bm E^{'''}
\end{align*}
where $\|\bm E^{''}\|=\mathcal{O}(\epsilon^2)$ and $\|\bm E^{'''}\|=\mathcal{O}(\epsilon^2)$. So, since $\bm{\mathrm{K}}$ and $\bm{\mathrm{\Lambda}}$ are diagonal matrices, if $B_{ij} = 0$, then
$(\bm \Gamma_\rho)_{ij} = \tau^{-1}_{ij}$ is also at most $\mathcal{O}(\epsilon^2)$. And if $B_{ij} \neq
0$, then the corresponding entry scales with $B_{ij}$.

The careful reader might wonder why we have constrained $\|\bm B\|<1/2$ instead of $<1$ thus far. This is to ensure that the Neumann series can be applied twice---for estimating $\bm \Sigma_\rho$ from $\bm \Gamma_\rho$, and $\bm\Gamma_\pi$ from $\bm \Sigma_\pi$. Since $\bm \Gamma_\rho = \bm I + \bm B$, setting $\|\bm B\| \le \delta < 1/2$ ensures that $\epsilon =  \frac{\delta}{1-\delta} <  1$ also. Along with the fact that $\lambda_i^2 \leq \kappa_i \,\, \forall i$, then the Neumann series can be applied a second time. However, for the rest of the paper we will relax this constraint slightly. While $\|\bm B\| < 1/2$ guarantees convergence of both the Neumann series, $\bm \Gamma_\rho$ matrices with higher operator norms of $\bm B$ (up to 1) could also allow for usable approximations, albeit with diminishing rates. This constraint on \(\bm B\) inherently limits the class of \(\bm \Gamma_\rho\) matrices compatible with our analysis; this is explored in greater detail in Section~\ref{ssec:share}. 
 
In the next section, we use these results to define a thresholding procedure to identify the undirected graphical model from the precision matrix.

\section{Learning Generalized Nonparanormal Independence Structure: An Algorithm}\label{sec:algo}

The theory above explains why precision matrices of the multivariate Gaussian ($\rho$) and the GNPN distribution ($\pi$) appear very similar. Large elements of $\bm \Gamma_\rho$ remain comparatively large in $\bm \Gamma_\pi$; zero elements of $\bm \Gamma_\rho$ are small in $\bm \Gamma_\pi$. By exploiting this dynamic, we can recover the conditional independence structure in the data, simply by computing the precision matrix of $\pi$ and thresholding it appropriately. As our algorithm uses a scalar threshold, we would need to compute the GNPN precision matrix from the correlation matrix (i.e. \(\bm R^{-1}_\pi\)), instead of computing it from \(\bm \Sigma_\pi\) (i.e. \( \bm \Sigma^{-1}_\pi \) ). In terms of Theorem~\ref{thm:main}, working with \(\bm R_\pi\) effectively means that matrices \(\bm K\) and \(\bm \Lambda\) are scaled with the appropriate variances. 

Below we list the variables (to help the reader distinguish them, as several have similar names), and then define a simple algorithm to recover the conditional independence structure by thresholding the GNPN precision matrix.

\begin{center}
\begin{table}[H]
    \centering
\begin{tabularx}{\textwidth}{||l X||} 
 \hline
 Variable& Definition \\
 \hline\hline
 $\bm \Gamma_\rho$& Precision matrix of the Gaussian distribution, scaled such that $(\bm \Gamma_\rho)_{ii} = 1 \,\, \forall i$ \\ 
 \hline
 $\bm B$& Non-zero off-diagonal entries of $\bm B$ correspond to edges, and $\bm \Gamma_\rho = \bm I + \bm B$\\
 \hline
 $\bm  \Sigma_\rho$& Covariance matrix of the Gaussian distribution\\
 \hline
 $\bm \Sigma_\pi$& Covariance matrix of the GNPN distribution\\
 \hline
 $\bm R_\pi$& Correlation matrix of the GNPN distribution\\
 \hline
 $\bm \Gamma_\pi$ & $:=\bm R_\pi^{-1}$, precision matrix of the GNPN distribution \\
 \hline
 $\bm \gamma^{\triangle}$ & Vector with magnitude of $\bm \gamma_{ij}$, entries in $\bm \Gamma_\pi$ where $i > j$, sorted in descending order\\
 \hline
 $t$ & Threshold \\
 \hline
 $\bm \Gamma_\pi^t$ & Thresholded $\bm \Gamma_\pi$; encodes conditional independence structure (if algorithm correct) \\
 \hline

\end{tabularx}
\caption{List of variables}
\label{table:variables}
\end{table}
\end{center}

Here, we give a step-by-step overview of Algorithm~\ref{alg:algo}.
\begin{enumerate}
    \item \textit{Compute Correlation:} From the GNPN data, compute the correlation matrix, \(\bm{R}_\pi\). Also compute \(\bm{\Gamma}_\pi\) as the inverse of \(\bm{R}_\pi\).
    \item \textit{Applicability Check:} To verify that this algorithm and analysis are applicable we would like to check if Neumann series approximation of matrix inverse can be applied twice---first to estimate $\bm  \Sigma_\rho$ from $\bm \Gamma_\rho$ and then to estimate $\bm  \Gamma_\pi$ from $\bm R_\pi$---as described previously. In practice, however, when presented with non-Gaussian data we would only be able to (easily) estimate $\bm  R_\pi$ and $\bm  \Gamma_\pi$, not $\bm  \Gamma_\rho$ and $\bm  \Sigma_\rho$. Hence, a necessary (but not sufficient) condition is to check if the Neumann series of $\bm R_\pi$ estimates well the empirically computed $\bm \Gamma_\pi$, for instance, by checking if $\|\bm R_\pi- \bm I\| < 1$. This applicability check is discussed in detail in Section~\ref{ssec:app_check}. 
    \item \textit{Obtaining Sorted Edge Weights:} Create $\bm \gamma^\triangle $ by extracting magnitudes of the strictly lower triangular entries from \(\bm \Gamma_\pi\) and sorting them in descending order.
    \item \textit{Threshold Identification:} Plot $\bm \gamma^\triangle$ and look for a sharp drop followed by flattening. This point corresponds to the transition from significant entries to those that are likely to be zero in $\bm \Gamma_\rho$. The threshold $t$ can be identified visually as an elbow or using the \textsc{Kneedle} algorithm ~\citep{satopaa2011finding}, which detects knees and elbows in data. \textsc{Kneedle} approximates the point of maximum curvature for discrete data. It does so by normalizing the input data, then identifying points that are local maxima in the difference between the value and index of a number. These maxima are potential elbow candidates. It then applies a criterion based on a user-defined sensitivity parameter to determine candidates that represent true elbows.  An implementation can be found in the Python package \texttt{kneed}. From our experiments, we recommend the following parameter values for the algorithm: \texttt{S = 1, curve = `convex', direction = `decreasing', online = True}. The \texttt{online = True} parameter allows \textsc{Kneedle} to continue its search for elbows after the initial one is found. Consequently it tends to correctly identify the threshold even when \(\bm \gamma^\triangle\) exhibits multiple plateaus, which could occur if edge weights are clustered around several distinct values.
    \item \textit{Conditional Independence Estimation:}  Set all the entries of $\bm \Gamma_\pi$ with magnitude less than or equal to the threshold $t$ to zero. This final step provides an estimate of the conditional independence properties as a thresholded precision matrix, $\bm\Gamma_\pi^t$, where a zero in the $ij$-th entry would represent conditional independence between variables $Z_i$ and $Z_j$, given a separator set in the graphical model.
\end{enumerate}

\begin{algorithm}
\caption{Estimating independence structure of GNPN data}
\label{alg:algo}
\KwData{samples $x \sim \pi$ }
\KwResult{thresholded precision matrix $\hat{\bm\Gamma}_\pi^t$}
Calculate empirical correlation $\hat{\bm R}_\pi$ and empirical precision $\hat{\bm \Gamma}_\pi$\;
\If{$\|\hat{\bm R}_\pi - \bm I\| \ge 1$}{
    Break \;
}
Extract the magnitudes of strictly lower triangular entries in \(\bm \Gamma_\pi\) and sort in descending order to obtain $\bm{\hat\gamma^\triangle}$\;
Identify a threshold value, $t$, by plotting $\bm{\hat\gamma^\triangle}$, or with \textsc{Kneedle}\;
Threshold $\hat{\bm\Gamma}_\pi$ with $t$ to get the thresholded precision matrix, $\hat{\bm\Gamma}_\pi^t$.
\end{algorithm}

\section{Example, Simulations and Applications}\label{sec:ex}

In this section, we demonstrate the proposed algorithm on synthetic experiments and some real-world data.

\subsection{Circular Graph Example}\label{ssec:circular}
Let us return to the example in Section~\ref{ssec:ex}. Recall the Gaussian precision is
$$\bm \Gamma_\rho = \left(
\begin{array}{cccccccc}
 1 & \alpha & 0 & 0 & 0 & 0 & 0 & \alpha \\
 \alpha & 1 & \alpha & 0 & 0 & 0 & 0 & 0 \\
 0 & \alpha & 1 & \alpha & 0 & 0 & 0 & 0 \\
 0 & 0 & \alpha & 1 & \alpha & 0 & 0 & 0 \\
 0 & 0 & 0 & \alpha & 1 & \alpha & 0 & 0 \\
 0 & 0 & 0 & 0 & \alpha & 1 & \alpha & 0 \\
 0 & 0 & 0 & 0 & 0 & \alpha & 1 & \alpha \\
 \alpha & 0 & 0 & 0 & 0 & 0 & \alpha & 1 \\
\end{array}
\right)$$
with $\alpha = 1/22 \approx 0.045$.

\begin{figure}[tbp]
    \centering
    \begin{subfigure}{0.49\textwidth}
        \centering
        \includegraphics[width=\linewidth]{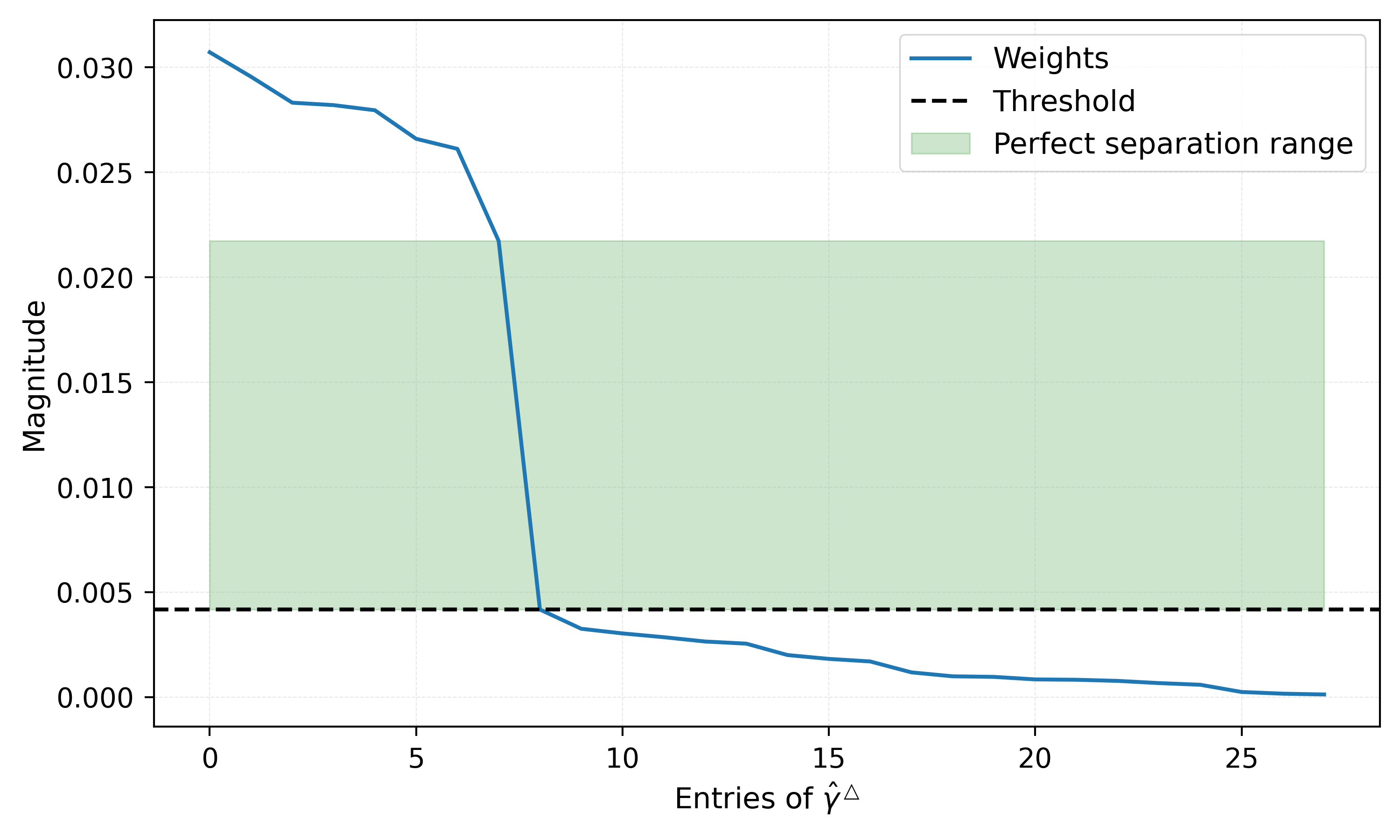}
        \caption{Correctly identified threshold}
        \label{fig:gamma_tri_plot_good}
    \end{subfigure}
    \hfill 
    \begin{subfigure}{0.49\textwidth}
        \centering
        \includegraphics[width=\linewidth]{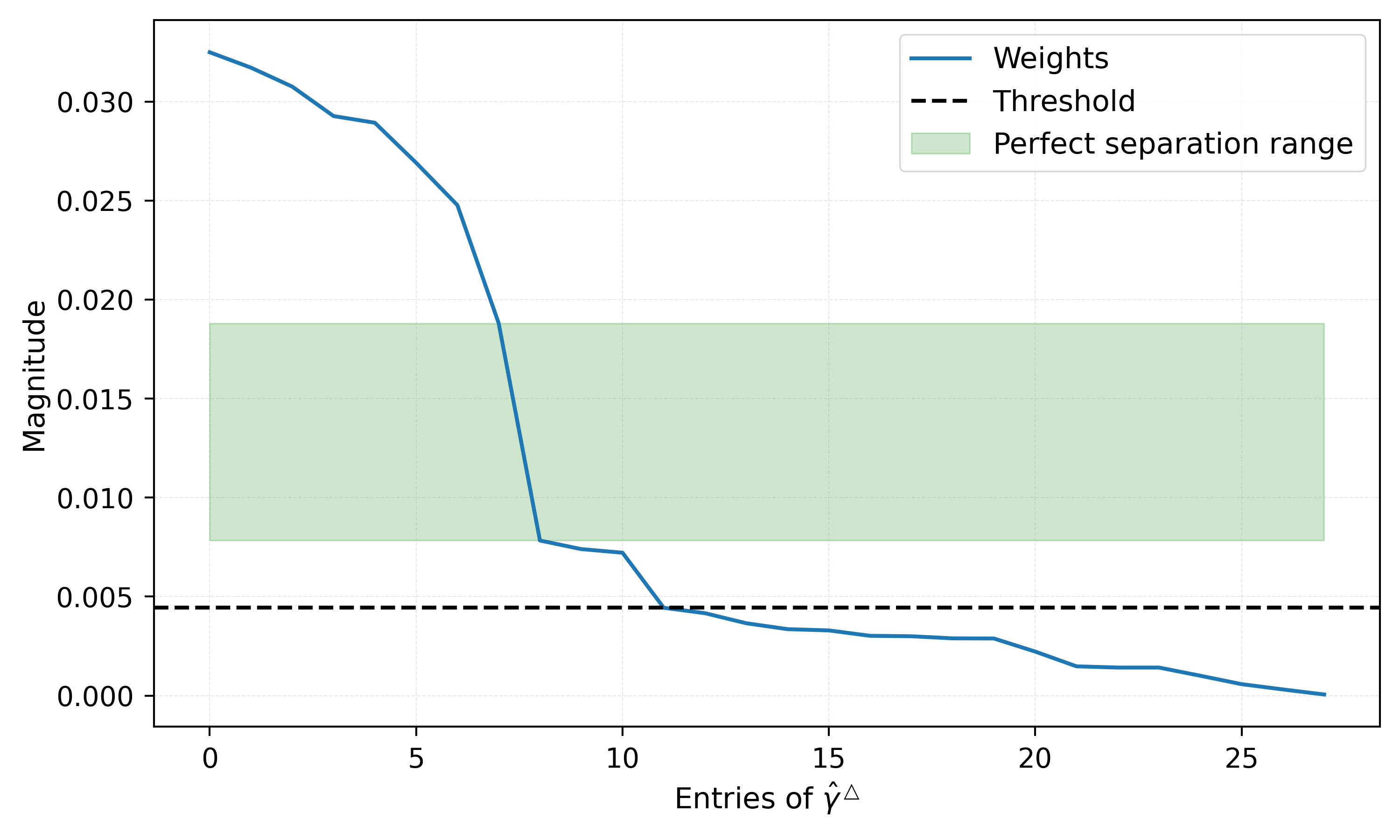}
        \caption{Incorrectly identified threshold}
        \label{fig:gamma_tri_plot_bad}
    \end{subfigure}
    \caption{Plot of \( \hat{\bm \gamma}^\triangle \), defined as a vector of edge magnitudes in \(\hat{\bm \Gamma}_\pi\) in descending order. The green bar represents the range where a threshold can be set to perfectly separate the edges and non-edges. Here the threshold has been identified by \textsc{Kneedle}. In the above two plots, \textsc{Kneedle} identifies a correct and an incorrect threshold for two different sets of data generated from the same initial $\bm\Gamma_\rho$.}
    \label{fig:gamma_tri_plot}
\end{figure}

First we generate 100,000 samples from $\rho = \mathcal{N}(\bm 0, \bm \Gamma_\rho^{-1} = \bm \Sigma_\rho)$ and transform them using $f_i(x)=x^3$ $\forall i$. For the normalized samples $\|\hat{\bm{R}}_\pi - \bm I\| < 1 $; hence we proceed with the rest of the algorithm. (The samples remain unnormalized in Figure~\ref{fig:gamma_tri_plot} and the displayed $\bm{\hat\Gamma_\pi^t}$ to be consistent with Section~\ref{ssec:ex}.) The threshold can be identified by plotting out $\hat{\bm \gamma}^\triangle$ to look for a point where it flattens. We use \textsc{Kneedle} to find the threshold at 0.0003 as marked in Figure~\ref{fig:gamma_tri_plot_good}.

Thresholding with this value gives us the following estimate of conditional independence matrix ($\bm{\hat\Gamma_\pi^t}$) with the same structure as $\bm\Gamma_\rho$.
$$ \bm{\hat\Gamma_\pi^t} = \left(
\begin{array}{cccccccc}
0.067 & 0.0019 & 0.0\,\,\,\,\, & 0.0\,\,\,\,\, & 0.0\,\,\,\,\, & 0.0\,\,\,\,\, & 0.0\,\,\,\,\, & 0.0017 \\ 0.0019 & 0.0666 & 0.0017 & 0.0\,\,\,\,\, & 0.0\,\,\,\,\, & 0.0\,\,\,\,\, & 0.0\,\,\,\,\, & 0.0\,\,\,\,\, \\ 0.0\,\,\,\,\, & 0.0017 & 0.0667 & 0.002 & 0.0\,\,\,\,\, & 0.0\,\,\,\,\, & 0.0\,\,\,\,\, & 0.0\,\,\,\,\, \\ 0.0\,\,\,\,\, & 0.0\,\,\,\,\, & 0.002 & 0.0655 & 0.0019 & 0.0\,\,\,\,\, & 0.0\,\,\,\,\, & 0.0\,\,\,\,\, \\ 0.0\,\,\,\,\, & 0.0\,\,\,\,\, & 0.0\,\,\,\,\, & 0.0019 & 0.067 & 0.0014 & 0.0\,\,\,\,\, & 0.0\,\,\,\,\, \\ 0.0\,\,\,\,\, & 0.0\,\,\,\,\, & 0.0\,\,\,\,\, & 0.0\,\,\,\,\, & 0.0014 & 0.0646 & 0.002 & 0.0\,\,\,\,\, \\ 0.0\,\,\,\,\, & 0.0\,\,\,\,\, & 0.0\,\,\,\,\, & 0.0\,\,\,\,\, & 0.0\,\,\,\,\, & 0.002 & 0.0654 & 0.0018 \\ 0.0017 & 0.0\,\,\,\,\, & 0.0\,\,\,\,\, & 0.0\,\,\,\,\, & 0.0\,\,\,\,\, & 0.0\,\,\,\,\, & 0.0018 & 0.0641 \\ 
\end{array}
\right)$$
The algorithm works perfectly on the above example. However, it is possible for it to fail to identify the correct threshold. This might happen because there in fact is no uniform threshold perfectly separating edges and non-edges, i.e., the smallest magnitude value in $\bm \Gamma_\pi$ corresponding to an edge in $\bm \Gamma_\rho$ is less than the largest magnitude value in $\bm \Gamma_\pi$ corresponding to a non-edge in $\bm \Gamma_\rho$. Another instance where the algorithm might identify the wrong threshold is when a knee in the plot of $\hat{\bm \gamma}^\triangle$ appears in the region where the values have already (mostly) flattened out. This behavior occurs with a different set of samples from the same initial precision matrix for the circular graph above: Figure~\ref{fig:gamma_tri_plot_bad} plots $\hat{\bm \gamma}^\triangle$ obtained from a new set of transformed samples, and \textsc{Kneedle} identifies a threshold lower than what it should. In this case, the algorithm would be unable to zero out edges in $\bm{\hat \Gamma}_\pi^t$ that are zero in $\bm \Gamma_\rho$. Note that here the algorithm, nevertheless, returned a graphical model consistent with the circular graphical model, but there are unnecessary edges: the algorithm returns an $\mathcal{I}$-map, but not the sparsest $\mathcal{I}$-map.

\subsection{Statistical Performance Experiment}\label{ssec:exp}

To estimate the statistical performance of Algorithm~\ref{alg:algo}, we run the experiment below a thousand times for each set of transformations listed in the first column of Table~\ref{table:1}.

Experiment procedure:~\label{exp:procedure}
\begin{enumerate}
  \item Randomly generate a $10 \times 10$ sparse Gaussian precision matrix, $\bm \Gamma_\rho$, with the Erdős–Rényi model with these steps:  \label{step:1}
  \begin{enumerate}
      \item Start with a zero matrix $\bm B$ \label{step:1a}
      \item Create each edge (off-diagonals) in $\bm B$ with probability $p$, where $p \sim \mathcal{U}[0.1,0.8]$
      \item For the edges present, assign an edge weight by randomly sampling from $\mathcal{N}(0, 0.3)$
      \item Set values in $\bm B$ with magnitude less than $0.1$ to $0$, to ensure we do not have arbitrarily small edge weights
      \item  Set the precision matrix $\bm \Gamma_\rho = \bm I + \bm B$
      \item Ensure positive-definiteness of \(\bm{\Gamma}_\rho\), else return to Step~\ref{step:1a}
      \item Check if \(\|\bm{B}\| < 1\); else, return to Step~\ref{step:1a} \label{step:1g}
   \end{enumerate}
  \item Generate $50,000$ samples with $\bm \Sigma_\rho = \bm \Gamma_\rho^{-1}$ \label{step:2}
  \item Transform each dimension of the samples $x_i$ with $f_i$:\label{step:3}
  \begin{enumerate}
      \item If using a single transformation function, transform all variables with it
      \item If using more than one transformation function, transform every variable with a randomly selected function from a given list
  \end{enumerate}
  \item Run Algorithm~\ref{alg:algo} to estimate the conditional independence structure, $\bm\Gamma_\pi^t$
\begin{enumerate}
      \item Check if $\| \bm R_\pi - \bm I\| < 1$; if not, return to Step~\ref{step:1a} \label{step:4a}
  \end{enumerate}
  \item Calculate the accuracy, precision and recall (for edges and non-edges in $\bm\Gamma_\pi^t$ as indicated by $\bm \Gamma_\rho$)
\end{enumerate}

\begin{center}
\begin{table}[tbp]
    \centering
\begin{tabular}{||c c c c||} 
 \hline
 Transformations & Accuracy (\%)& Recall (\%)& Precision (\%)\\
 \hline\hline
 $\sin(x)$& 96.0 (6.8)& 97.1 (10.8)& 85.7 (16.2)\\ 
 \hline
 $\cos(x)$& 93.2 (9.5)& 83.0 (18.8)& 89.9 (18.0)\\
 \hline
 $x^2$ & 95.8 (7.6)& 87.0 (16.6)& 95.5 (14.2)\\
 \hline
 $x^3$ & 98.2 (2.4)& 99.7 (3.0)& 90.5 (12.6)\\
 \hline
 $x^7$& 90.3 (7.8)& 77.3 (22.1)&80.4 (20.8)\\
 \hline
 $x^3-x^2$ & 97.2 (3.2)& 99.4 (4.1)& 86.4 (14.0)\\
 \hline
 Power transformation & 98.2 (2.7)& 99.7 (3.1)& 90.6 (13.0)\\
 \hline
 CDF transformation & 99.1 (1.6)& 100 (0.6)& 94.5 (9.8)\\
 \hline
 $[\sin(x), x^3]$ & 93.9 (6.6)& 98.0 (8.7)&78.6 (17.1)\\
 \hline
 $[\sin(x), \cos(x)]$& 80.1 (14.7)& 43.0 (19.0)&63.3 (26.7)\\
 \hline
 $\sin(2x)$& 83.0 (14.5)& 77.9 (22.2)&65.5 (23.8)\\
 \hline
\end{tabular}
\caption{Mean (std) accuracy, recall, and precision of conditional independence structure recovered by Algorithm~\ref{alg:algo} from synthetic GNPN data generated with the Erdős–Rényi graphical model with a random percolation probability.}
\label{table:1}
\end{table}
\end{center}

The power and CDF transformations listed in Table~\ref{table:1}, defined in~\citet{liu2009nonparanormal}, are shown below for completeness.

\textit{Power Transformation}: Let $f_0(t)$ be a symmetric and odd transformation given by $f_0(t) = \text{sign}(t)|t|^{\alpha}$, where $\alpha > 0$ is a parameter. The power transformation for the $j$-th dimension is:
$$f_j(z_j) \equiv \sigma_j \left( \frac{f_0(z_j-\mu_j)}{\sqrt{\int f_0^2(t-\mu_j)  \phi\left(\frac{t-\mu_j}{\sigma_j} \right)dt}} \right) + \mu_j,$$
where $\mu_j$ and $\sigma_j$ are the mean and standard deviation of the $j$-th dimension, with $\sigma_j = \Sigma_0(j,j)$. As in~\cite{liu2009nonparanormal}, for the experiments we used $\alpha=3$.

\textit{Gaussian CDF Transformation}: Let $f_0(t)$ be a one-dimensional Gaussian cumulative distribution function (CDF) with mean $\mu_{f_0}$ and standard deviation $\sigma_{f_0}$, such that $f_0(t) = \Phi\left(\frac{t-\mu_{f_0}}{\sigma_{f_0}}\right)$. The transformation function $f_j$ for the $j$-th dimension is:
$$f_j(z_j) \equiv \sigma_j \left( \frac{f_0(z_j) - \int f_0(t)\phi\left(\frac{t-\mu_j}{\sigma_j}\right)dt}{\sqrt{\int\left(f_0(y) - \int f_0(t)\phi\left(\frac{t-\mu_j}{\sigma_j}\right)dt\right)^2\phi\left(\frac{y-\mu_j}{\sigma_j}\right)dy}} \right) + \mu_j,$$
where $\phi$ is the standard normal probability density function. For the experiments the parameters used were the same as in~\cite{liu2009nonparanormal} with $\mu_{f_0}=0.05$ and $\sigma_{f_0}=0.4$.

\begin{figure}[tbp]
    \centering
    \begin{subfigure}{0.32\textwidth}
        \centering
        \includegraphics[width=\linewidth]{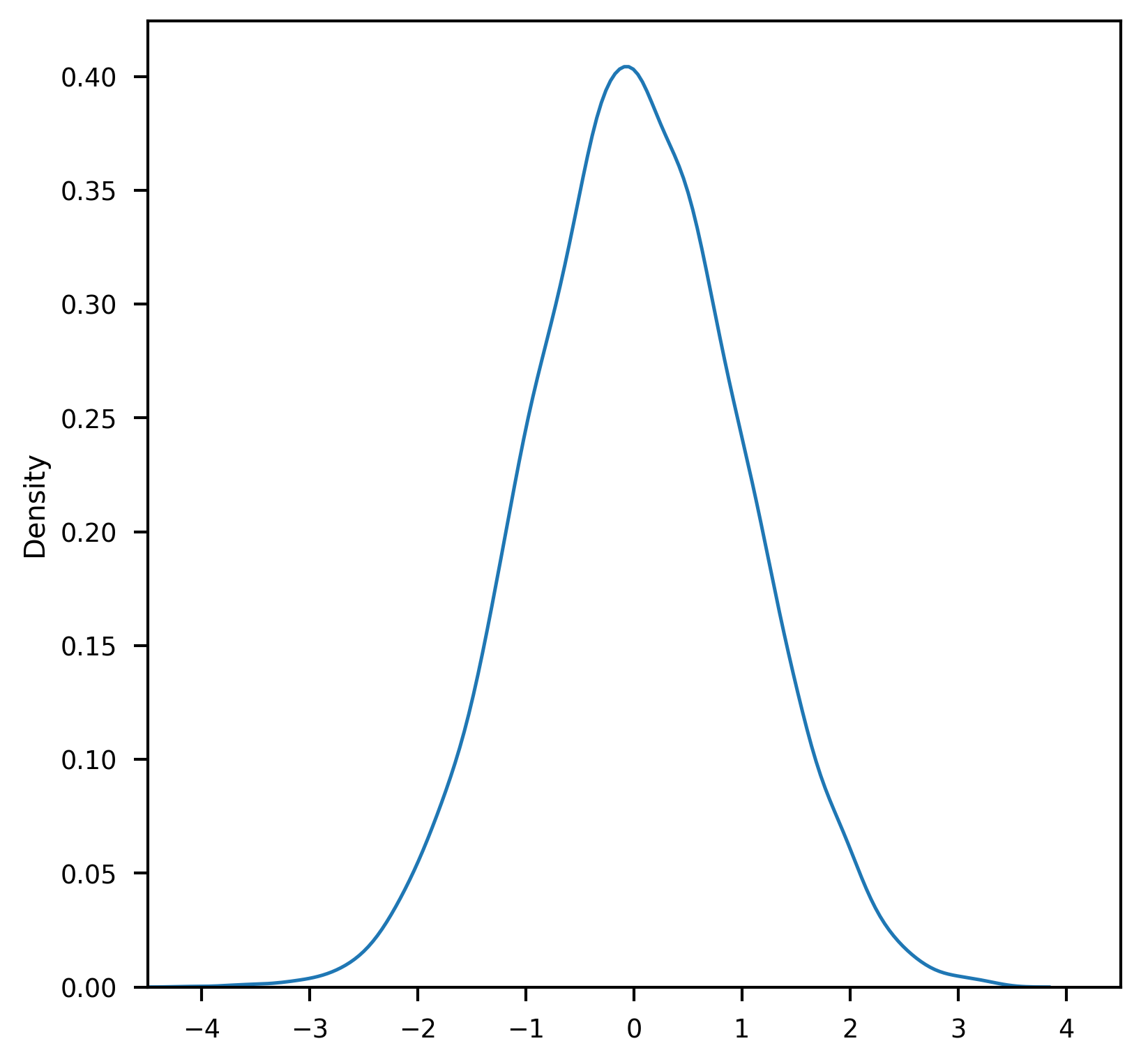}
        \caption{Standard normal}
        \label{fig:std_normal}
    \end{subfigure}
    \hfill
    \begin{subfigure}{0.32\textwidth}
        \centering
        \includegraphics[width=\linewidth]{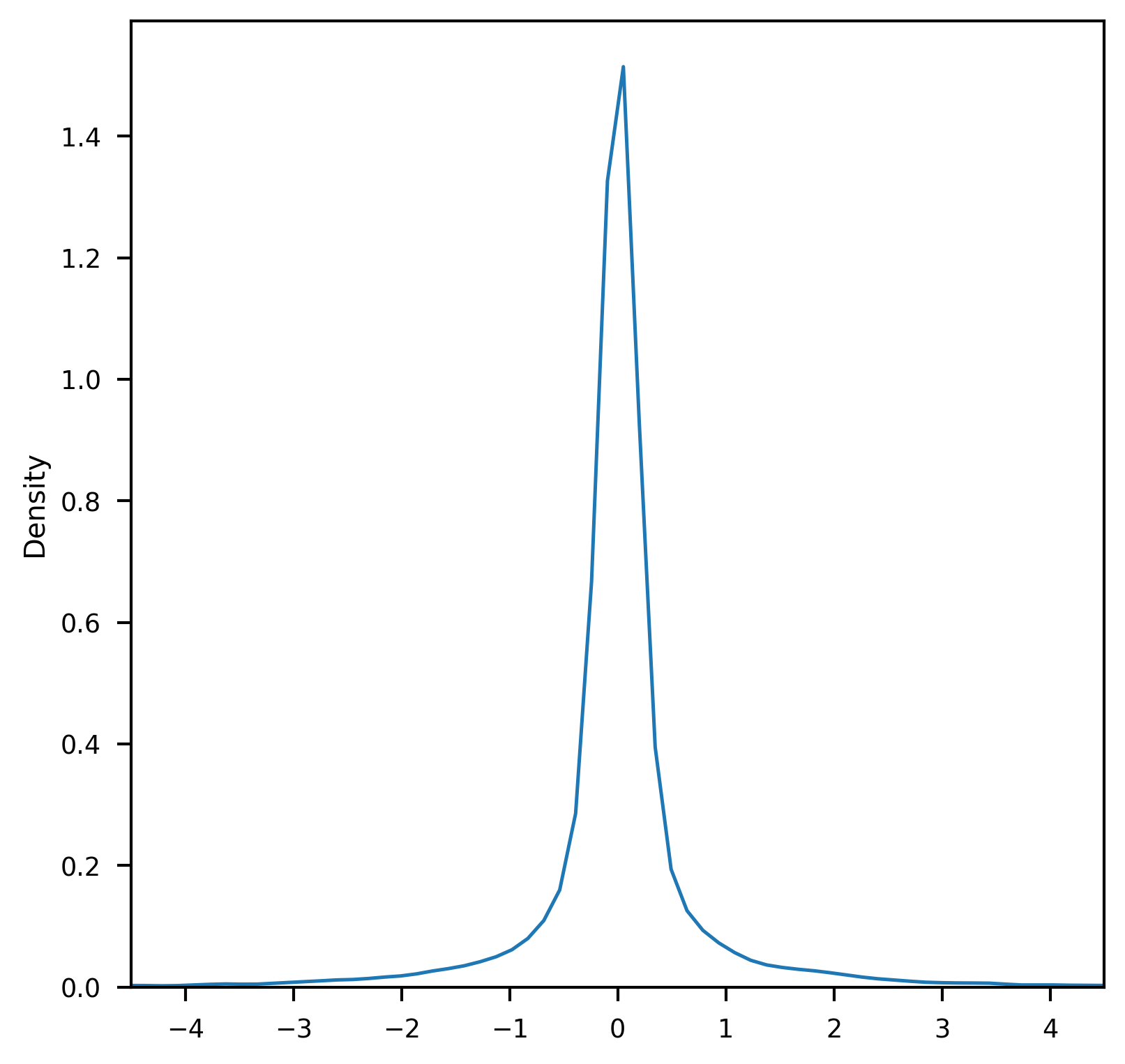}
        \caption{Power transformation}
        \label{fig:power}
    \end{subfigure}
    \hfill
    \begin{subfigure}{0.32\textwidth}
        \centering
        \includegraphics[width=\linewidth]{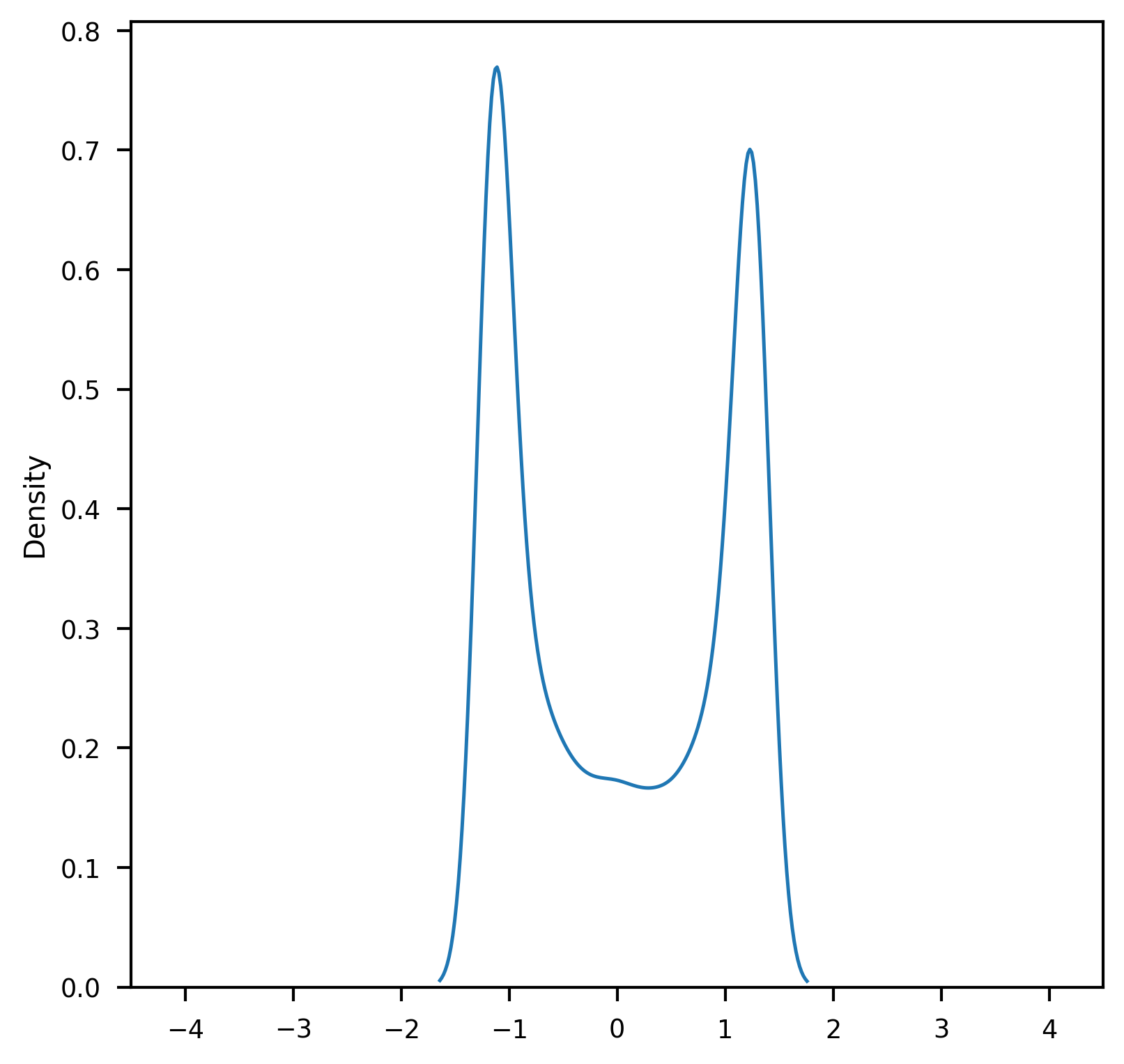}
        \caption{CDF transformation}
        \label{fig:CDF}
    \end{subfigure}
    \caption{Density plot of a standard normal distribution before and after application of power and CDF transformations.}
    \label{fig:power_cdf_trans}
\end{figure}

The two transformations are visualized in Figure~\ref{fig:power_cdf_trans}, by applying them to samples from a standard normal distribution. The power transformations maps the normal distribution into a sharply peaked distribution, while the CDF transformation maps it closer to a bimodal distribution.

For different transformations, Table~\ref{table:1} reports the mean accuracy for both edges and non-edges, along with recall and precision specifically for edges. The experiments demonstrate that, for many transformations, the proposed algorithm recovers the graph structure with high accuracy, generally performing best for odd functions, as expected from the theory in Sections~\ref{sec:cov} and~\ref{sec:est_gamma}.

Even transformations do not preserve signs, compressing edge weights in $\bm \Gamma_\pi$ towards smaller values and making it challenging for the algorithm to differentiate true dependencies from noise. This leads to a higher likelihood of missing some true conditional dependencies, resulting in lower recall. But this compression also reduces the occurrence of spurious dependencies, leading to higher precision in the recovered structure. 

Higher-order polynomials, such as $x^7$, also pose a challenge. They compress weights in $\bm \Gamma_\pi$ to close to zero (making it harder to differentiate true edges from weights), but also introduce outliers that would dominate the empirical precision calculation (introducing spurious dependencies). This results in both low precision and low recall for these types of transformations.

The algorithm also performs poorly for functions like $\sin(2x)$ that do not have uniformly bounded derivatives (although they still satisfy condition~\ref{der.est}). The theory in Section~\ref{sec:est_gamma}, especially Lemma~\ref{tau:est}, shows that the estimates suffer when the derivatives are allowed to grow. Experiments also show that when some transformation functions are even while others odd leads to poor results (as shown by the pair $(\sin(x), \cos(x))$. This is expected as transforming different variables with purely odd and even functions destroys any dependence between them. To be more specific, if $f_i$ is even and $f_j$ is odd, then $\tau_{ij} = 0$ since $\tau_{ij}$ is computed as the inner product between two orthogonal functions.
 
\subsection{Experiments with the Galton-Watson Process Trees}

We now further analyze the performance of the algorithm on branching processes or tree structures. We use the Galton-Watson process \citep{watson1875probability} as a way to generate random trees owing to its simplicity and because of the branching behavior found in many natural systems. Each tree is created by assigning a Poisson distributed (\(\lambda = 2\)) number of children to the single root node, and recursively assigning children similarly until a prespecified number of nodes (10) is reached. Each edge (from parent to child) is assigned a weight by randomly sampling from $\mathcal{N}(0, 0.3)$, while also enforcing a minimum absolute value edge weight of 0.1 like before. We also make sure that the Neumann series can be used to approximate the matrix inverse of $\bm \Gamma_\rho$ and $\bm R_\pi$. 50,000 random Gaussian samples are generated per tree with independence information as given by the tree. The samples are then transformed to generalized nonparanormal form, from which we try to recover the conditional independence structure. We conduct this experiment a thousand times for each transformation. The average performance of Algorithm~\ref{alg:algo} is shown in Table~\ref{table:galton_watson_results}. Comparing them with the results in Table~\ref{table:1}, we see that the performance metrics remain similar to the case of generated random graphs with the Erdős–Rényi model.

\begin{center}
\begin{table}[tbp]
    \centering
\begin{tabular}{||c c c c||} 
 \hline
 Transformations & Accuracy (\%)& Recall (\%)& Precision (\%)\\
 \hline\hline
 $\sin(x)$& 95.0 (7.8)& 96.3 (13.5)& 84.8 (18.0)\\ 
 \hline
 $\cos(x)$& 93.5 (7.9)& 83.4 (18.1)& 88.0 (19.0)\\
 \hline
 $x^2$ & 97.9 (3.0)& 90.1 (14.8)& 99.6 (2.3)\\
 \hline
 $x^3$ & 98.5 (2.7)& 99.6 (4.3)& 94.4 (8.8)\\
 \hline
 $x^7$ & 91.5 (4.0)& 73.3 (20.8)& 85.9 (12.9)\\
 \hline
 $x^3-x^2$ & 97.3 (4.1)& 99.7 (3.0)& 90.5 (11.8)\\
 \hline
 Power transformation & 98.5 (2.3)& 99.8 (2.3)& 94.2 (8.3)\\
 \hline
 CDF transformation & 99.2 (1.5)& 100 (0.0)& 96.6 (6.1)\\
 \hline
 $[\sin(x), x^3]$& 91.1 (7.8)& 97.9 (8.3)&74.0 (17.8)\\
 \hline
\end{tabular}
\caption{Mean (std) accuracy, recall and precision of conditional independence structure recovered from generalized nonparanormal data sampled from a Galton-Watson process tree.}
\label{table:galton_watson_results}
\end{table}
\end{center}

In the following two subsections, we evaluate the proposed algorithm on two datasets: one where our method performs robustly despite challenging conditions and another where performance degrades owing to significant deviations from the algorithm's theoretical requirements.

\subsection{Rochdale Data} \label{ssec:rochdale}
The Rochdale data set, a social survey analyzed in~\cite{whittaker1990graphical}, is an observational study conducted in Rochdale, UK, aimed to identify factors influencing women's economic activity. The dataset comprises 655 observations across eight binary variables:

\begin{enumerate}[label=\alph*]
    \item : Wife economically active (no, yes)
    \item : Age of wife \(>38\)  (no, yes)
    \item : Husband unemployed (no, yes)
    \item : Child \(\le 4\) (no, yes)
    \item : Wife's education, high-school+ (no, yes)
    \item : Husband's education, high-school+ (no, yes)
    \item : Asian origin (no, yes)
    \item : Other household member working (no, yes)
\end{enumerate}

Previous work by one of the authors \citep{morrison2024exact} provides the theoretical framework for exact covariance transformation for $L^2$-functions through Fourier and Laplace transforms. This allows us to apply Algorithm~\ref{alg:algo} to this data set even as all the variables are binary. Nevertheless, the Rochdale data set challenges our algorithm in several aspects. Some variables (such as ``Age of wife") can be thought of as binary transformations of underlying Gaussian variables. However, the step functions used to binarize the data would need to have compact support to be in $L^2$---we meet this by assuming the right-side support to be sufficiently large. Further, some variables (such as ``Asian origin") are inherently categorical, violating our assumption that the pre-transformed data follows a Gaussian distribution. Finally, the correlation matrix obtained from the data is slightly above the threshold to estimate the inverse through Neumann series, with \(\|\bm{R}_\pi-\bm{I}\| = 1.13\).  Despite these theoretical incompatibilities, we proceed with this dataset as a stress test for our method. 

\begin{figure}[tbp]
    \centering
    \begin{subfigure}{0.32\textwidth}
        \centering
        \includegraphics[width=\linewidth]{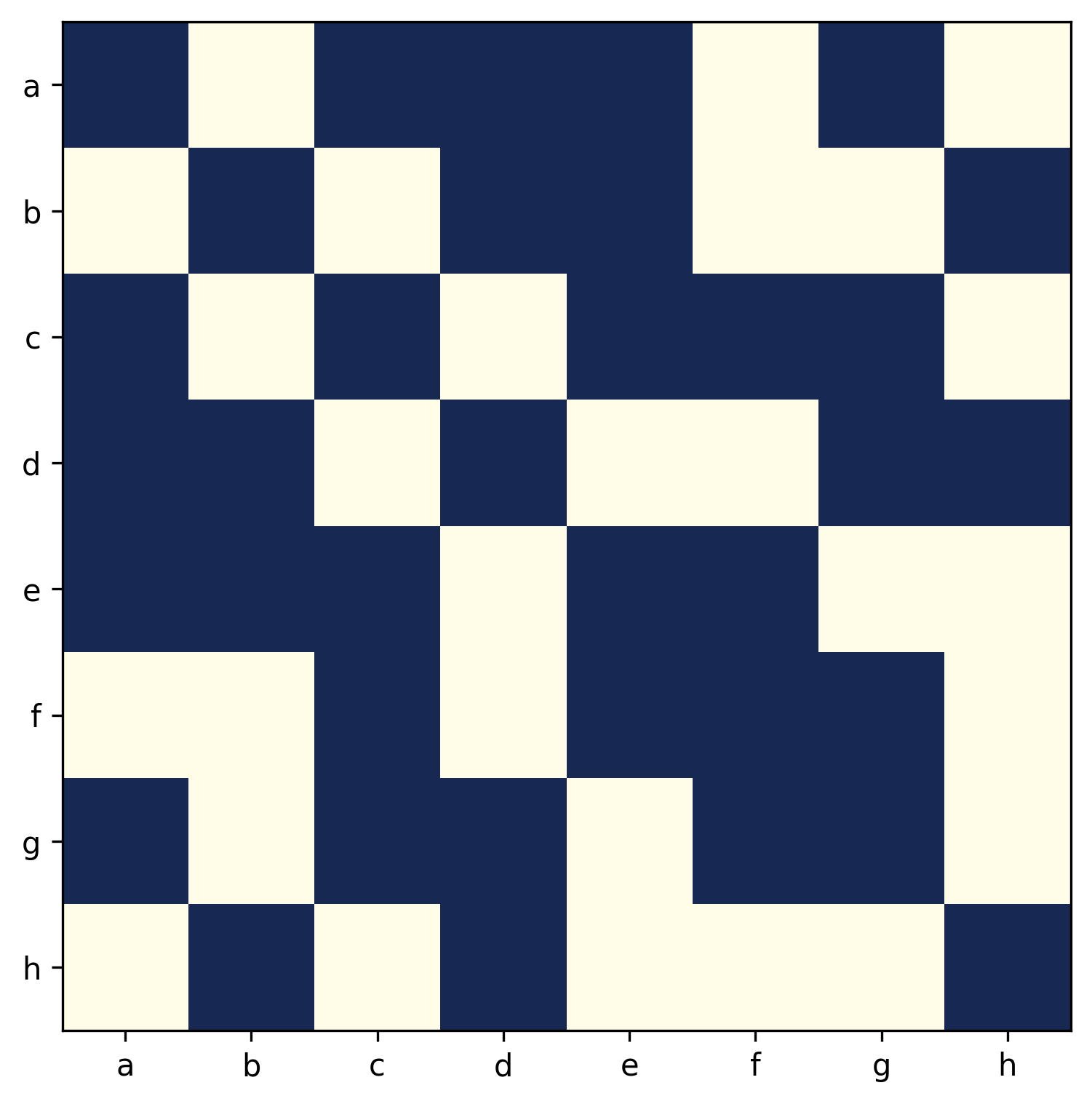}
        \caption{Reference algorithm}
        \label{fig:rochdale_reference}
    \end{subfigure}
    \hfill
    \begin{subfigure}{0.32\textwidth}
        \centering
        \includegraphics[width=\linewidth]{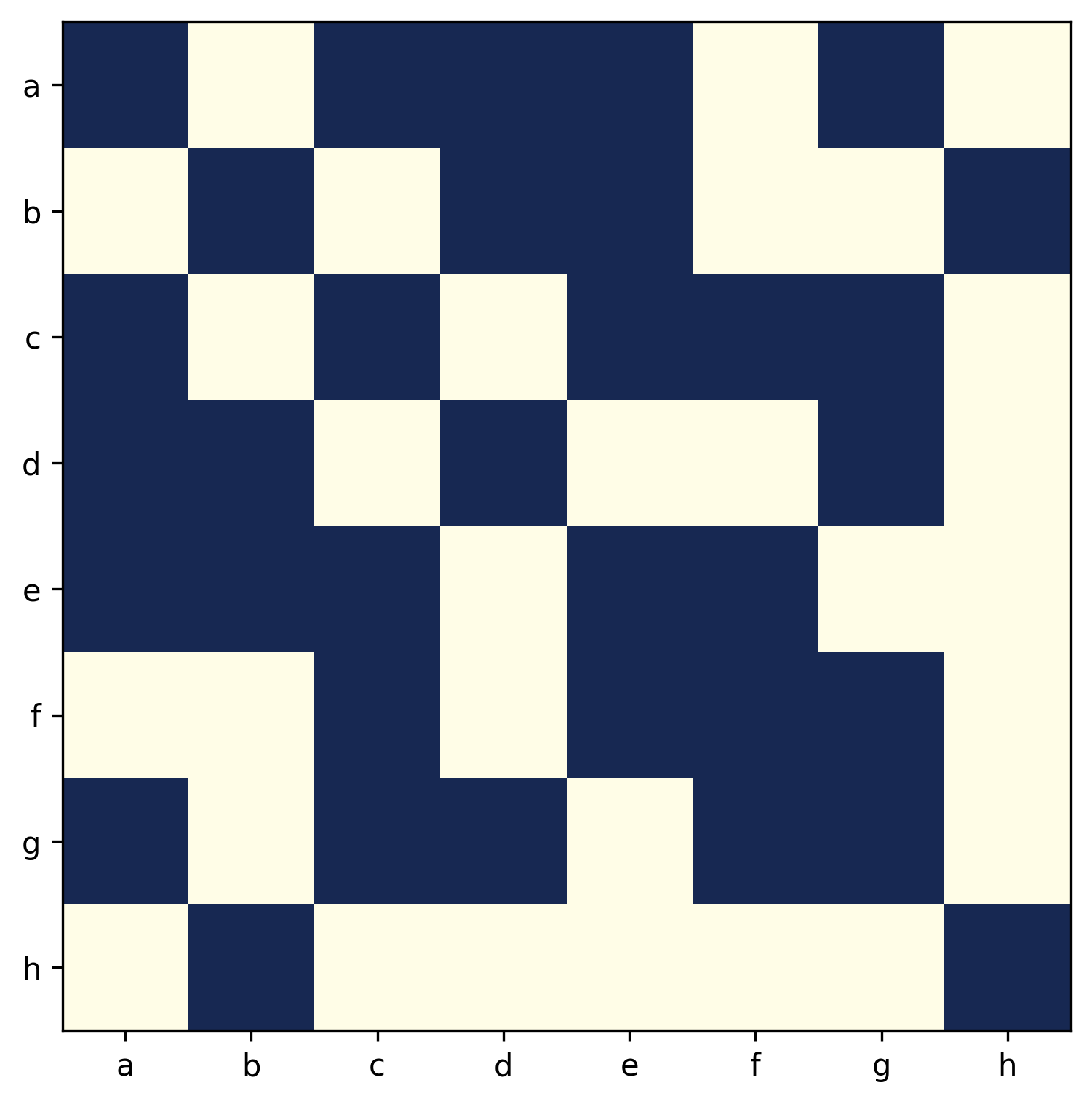}
        \caption{Algorithm~\ref{alg:algo}}
        \label{fig:rochdale_gnpn}
    \end{subfigure}
    \hfill
    \begin{subfigure}{0.32\textwidth}
        \centering
        \includegraphics[width=\linewidth]{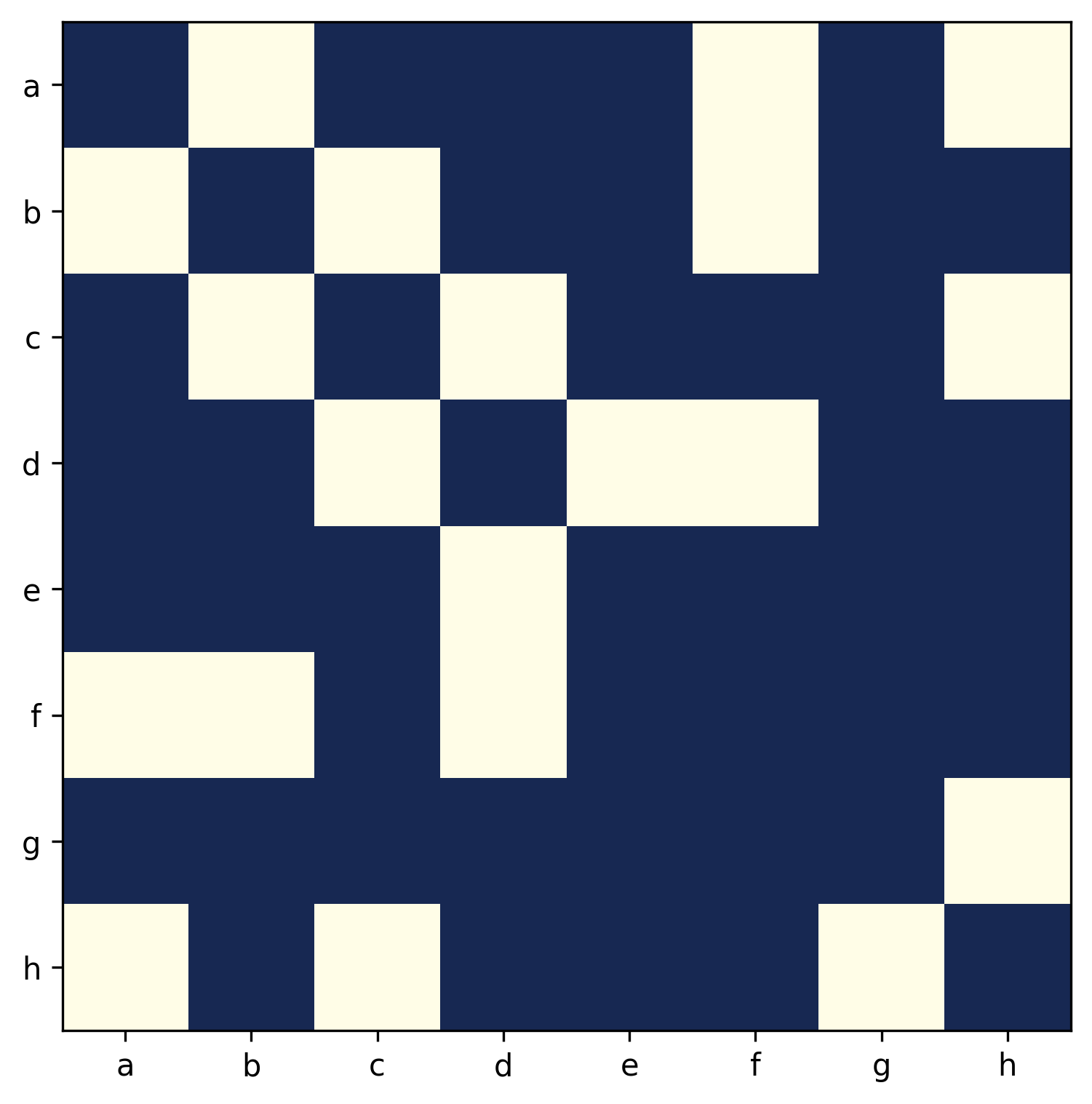}
        \caption{Nonparanormal SKEPTIC}
        \label{fig:rochdale_skeptic}
    \end{subfigure}
    \caption{Adjacency matrices of the graphs learned by different algorithms for the Rochdale data set. The graph learned by the Algorithm~\ref{alg:algo} is nearly identical to the reference described in~\protect\cite{whittaker1990graphical}. The nonparanormal SKEPTIC algorithm is described in~\protect\cite{liu2012nonparanormal}. }
    \label{fig:rochdale_graphs}
\end{figure}

Following Algorithm~\ref{alg:algo} and using \textsc{Kneedle} to determine the threshold, our approach recovers the graphical model shown in Figure~\ref{fig:rochdale_gnpn} closely aligning with the relationships modeled by~\cite{whittaker1990graphical} (Figure~\ref{fig:rochdale_reference}), which we use as the reference graph. The algorithm identifies all but one of the edges and non-edges correctly. Specifically, the graph identifies 13 of the 14 relationships validated by the original analysis, missing the edge \((d,h)=(h,d)\). The reference graph and the one learned by Algorithm~\ref{alg:algo} agree on the neighborhood of variable \(a\) (wife's economic activity) as \(c\) (husband's unemployment), \(d\) (child \(\le 4\)), \(e\) (wife's education, high-school+) and \(g\) (Asian origin). 

For comparison, we also learned a graphical model for this dataset using the nonparanormal SKEPTIC algorithm described in~\cite{liu2012nonparanormal}. While this algorithm is not designed to handle discontinuous generalized nonparanormal data, we include it owing to its simplicity and performance on nonparanormal data. Our implementation used Spearman's rank correlation to estimate a correlation matrix, subsequently inverted using graphical lasso (GLASSO)~\citep{friedman2008sparse} to obtain the estimated conditional independence structure. Using Stability Approach to Regularization Selection (StARS)~\citep{liu2010stability} to tune the penalty parameter in GLASSO, the technique recovers a different conditional independence structure compared to the reference. However, we were able to recover the correct structure with prior knowledge of the true graph and manual adjustment of the penalty parameter. This highlights the advantage of our approach, which did not require parameter tuning to achieve accurate results on non-Gaussian Rochdale data.

\subsection{Cell Signaling Data}\label{ssec:sachs}

\begin{figure}[tbp]
    \centering
    \begin{subfigure}{0.32\textwidth}
        \centering
        \includegraphics[width=\linewidth]{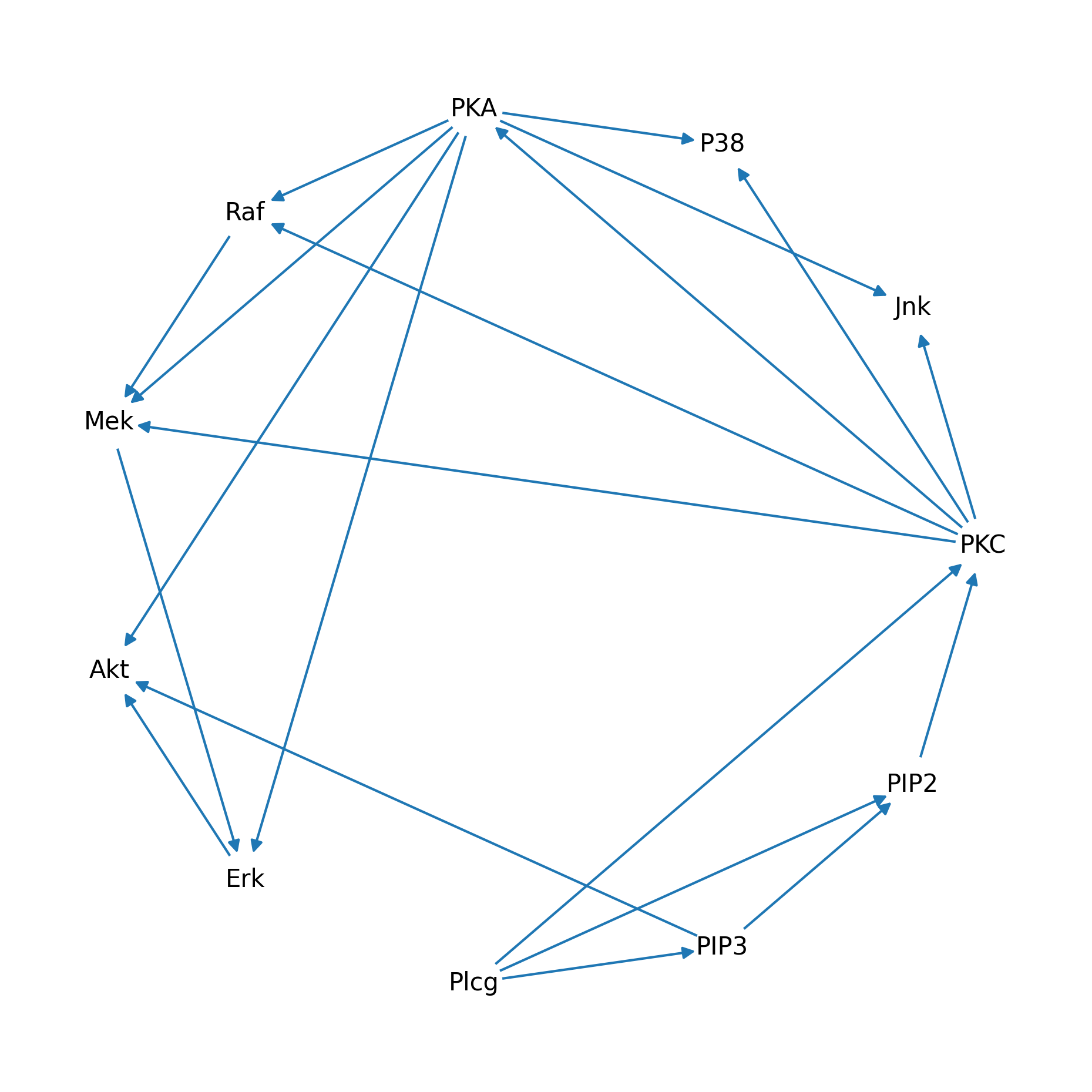}
        \caption{Reference model}
        \label{fig:sachs_reference}
    \end{subfigure}
    \hfill
    \begin{subfigure}{0.32\textwidth}
        \centering
        \includegraphics[width=\linewidth]{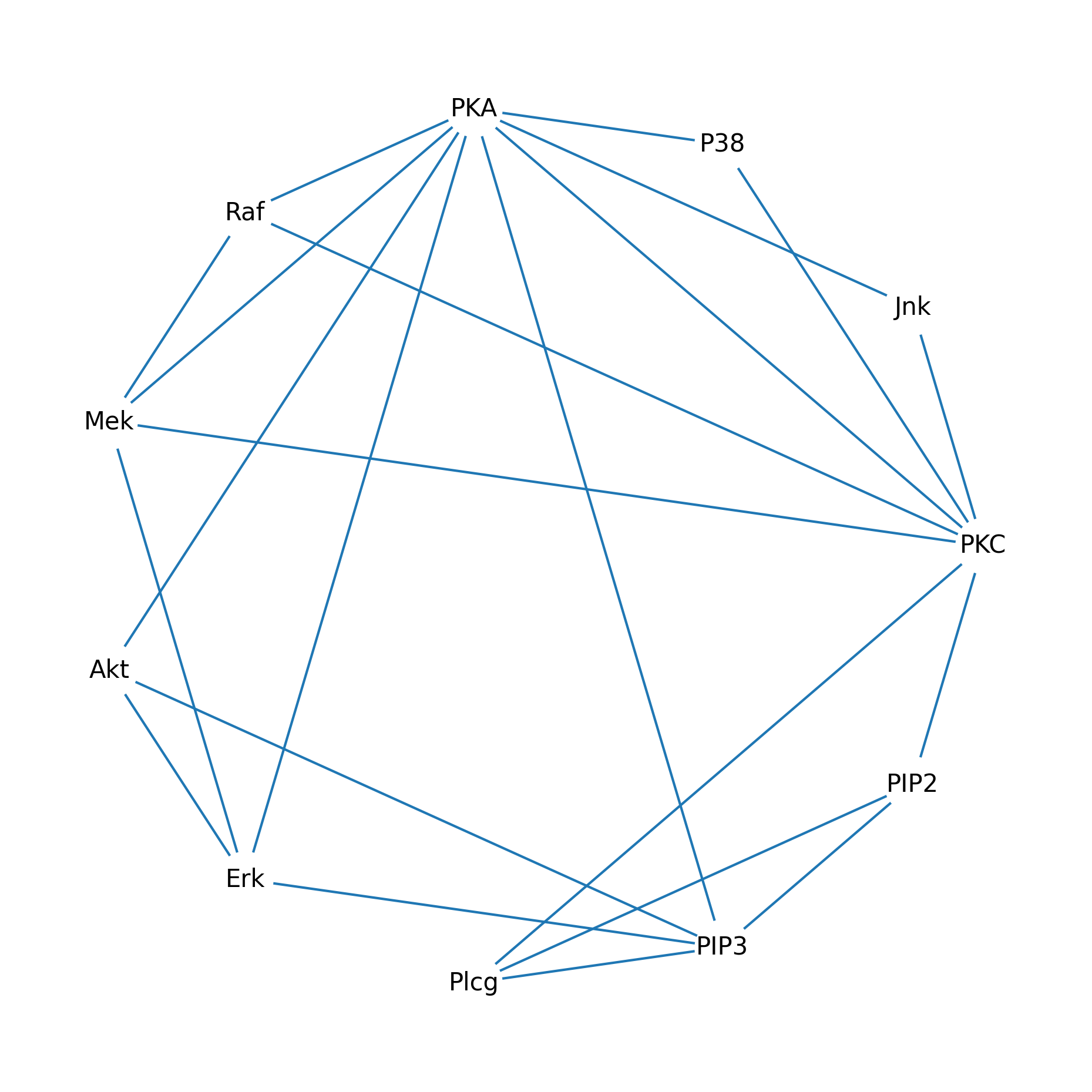}
        \caption{Moralized reference model}
        \label{fig:sachs_ref_moral}
    \end{subfigure}
    \hfill
    \begin{subfigure}{0.32\textwidth}
        \centering
        \includegraphics[width=\linewidth]{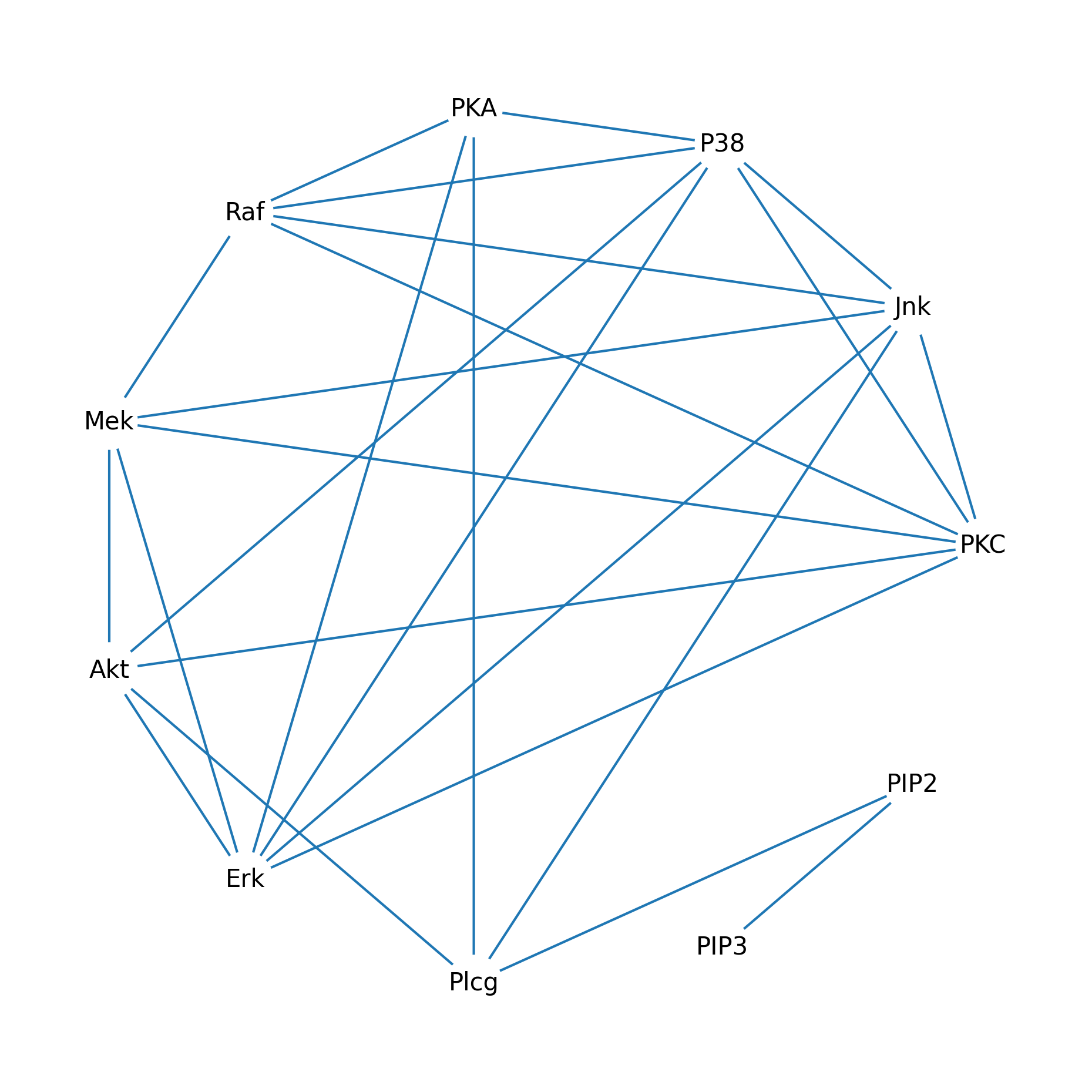}
        \caption{Algorithm~\ref{alg:algo} output}
        \label{fig:sachs_gnpn}
    \end{subfigure}
     \caption{Graphical models for a cell-signaling data set~\protect\citep{sachs2005causal}. The relationships in the reference (a) were verified experimentally. Our proposed algorithm learns a sub-optimal model, as compared with the moralized reference (b), as the data does not meet its requirements.}
    \label{fig:sachs_graphs}
\end{figure}

We now apply our algorithm to infer the conditional dependence structure for a cellular signaling network, based on single-cell data obtained via flow cytometry~\citep{sachs2005causal}. The data set contains \(n=7446\) simultaneous measurements of \(d = 11\) phosphorylated proteins and phospholipids derived from thousands of individual primary immune system cells, subjected to both general and specific molecular interventions. This data set has been widely used in prior studies, including~\cite{friedman2008sparse} and~\cite{baptista2024learning}, which applied GLASSO and the SING algorithm, respectively, to estimate the underlying graphical model. As the reference model, we use the directed graph identified in~\cite{sachs2005causal}, displayed here in Figure~\ref{fig:sachs_reference}, where relationships were verified experimentally. 

Each variable in the data exhibits a strong right skewness, with a large concentration of values near zero and a long positive tail. As the measurements are all non-negative, we first apply a marginal log transformation to each variable to make the data more suitable to the algorithms (this diagonal transformation does not change the graph structure). Importantly, however, the transformed data fails to satisfy the spectral norm condition on the correlation matrix required by our method. We observe \(\|\bm{R}_\pi - \bm{I}\| = 3.73\), significantly exceeding our threshold of 1. Consequently, we anticipate degraded performance of our algorithm in recovering the conditional dependence structure.

As with the preceding example, we identify the threshold with \textsc{Kneedle}. Of the 55 edges and non-edges in the complete graph, the graph inferred by the algorithm correctly identifies 32 edges and non-edges from the moralized reference graph while misclassifying 23 edges and non-edges (13 FPs and 10 FNs). This performance is similar to that of GLASSO (23 misclassifications) and slightly better than the nonparanormal SKEPTIC (26 misclassifications). The nonparanormal SKEPTIC algorithm was implemented as described previously, except that the optimal penalty parameter was chosen through 5-fold cross-validation. Notably, the SING algorithm achieves superior accuracy, correctly identifying 37 edges and non-edges, with only 18 misclassifications (12 FPs and 6 FNs). This improved performance is attributable to its design, which accommodates more general non-Gaussian data. 

\section{Practical Considerations} \label{sec:prac_considerations}

We now discuss several practical considerations that arise when applying our proposed algorithm. First, we address the accuracy of the applicability check in ensuring the validity of our theoretical assumptions. Next, we investigate how frequently typical precision matrices satisfy these assumptions in practice. Finally, we analyze the sample efficiency of our algorithm. 

\subsection{Applicability Check for Neumann Inverse Assumptions}\label{ssec:app_check}
Apart from assuming that the distribution is of the generalized nonparanormal type, perhaps the main limitation with our algorithm is that it assumes Neumann series can be used to approximate the inverses of $\bm \Gamma_\rho$ and $\bm R_\pi$. Specifically, we require $\|\bm R_\pi - \bm I\|<1$ as an applicability check.  Although this heuristic does not guarantee that $\bm \Sigma_\rho$ is estimable through the Neumann inverse of $\bm \Gamma_\rho$, to the best of our knowledge, it remains the only feasible approach short of explicitly estimating the transformations $g_i$ associated with the generalized nonparanormal distribution.

\begin{center}
\begin{table}[tbp]
    \centering
\begin{tabular}{||c || c || c c c ||}
 \hline
 Transformations & \% falsely passing& Accuracy (\%)&  Recall (\%)& Precision (\%)\\
 &applicability check&&&\\
 \hline\hline
 $\sin(x)$& 7.7& 94.7 (9.1)& 94.8 (15.3)& 84.8 (16.9)\\ 
 \hline
 $\cos(x)$& 12.1&  91.5 (10.2)& 79.4 (20.4)& 89.2 (17.1)\\
 \hline
 $x^2$ & 3.8& 95.8 (5.6)& 86.1 (16.4)& 95.7 (12.6)\\
 \hline
 $x^3$ & 1.1& 98.0 (3.0)& 99.6 (4.1)& 90.0 (13.3)\\
 \hline
 $x^7$& 9.1& 89.3 (8.5)& 75.6 (23.2)&80.5 (20.6)\\
 \hline
 $x^3-x^2$ & 1.5& 97.3 (3.3)& 99.2 (4.4)& 87.6 (13.8)\\
 \hline
 Power transformation & 1.8& 98.1 (2.8)& 99.7 (2.8)& 90.7 (12.6)\\
 \hline
 CDF transformation & 0.9& 99.0 (1.7)& 100 (0.7)& 94.1 (9.7)\\
 \hline
 $[\sin(x), x^3]$& 4.6& 93.1 (8.0)& 96.6 (11.2)& 78.6 (16.8)\\
 \hline
\end{tabular}
\caption{Checking if $\|\bm R_\pi - \bm I\| < 1$ generally performs well as a way to ensure that the algorithm is applicable.  Accuracy, recall and precision are means (std) for a thousand experiments that pass the applicability test, regardless of whether $\|\bm B\| < 1$.}
\label{table:app_check}
\end{table}
\end{center}

Table~\ref{table:app_check} shows that this applicability check generally performs well as a way to test whether Neumann series can approximate the inverses of $\bm\Gamma_\rho$ and $\bm R_\pi$. This is also supported by the accuracy, recall and precision metrics for model recovery observed across all experiments that pass the check, including cases where $\bm\Sigma_\rho$ cannot be approximated with the Neumann series of $\bm \Gamma_\rho$. Results in Table~\ref{table:app_check} were obtained using the same methodology as in the experiment conducted in Section~\ref{exp:procedure}, except for Step~\ref{step:1g} as this step would not be directly verifiable when given GNPN data alone.

As also seen in the table, the effectiveness of the check varies across different transformations. The highest proportion of false positives---where $\|\bm R_\pi - \bm I \| < 1$ but $\bm\Sigma_\rho$ is not estimable through the Neumann series of $\bm \Gamma_\rho$---occurs with the $\sin(x)$, $\cos(x)$, and $x^7$ transformations. This may be due to the cyclic nature of sine and cosine, and the high non-linearity of $x^7$, which disrupt the relationship between the independence structures encoded in $\bm\Sigma_\rho$ and $\bm R_\pi$. For these functions, variables that are correlated (non-zero values in $\bm \Sigma_\rho$) often become nearly uncorrelated after the transformation (close to zero values in $\bm R_\pi$). Consequently, the operator norm of $\bm R_\pi$ decreases sufficiently for its inverse to be approximated by the Neumann series, even in cases where \(\Vert \bm B \Vert \ge 1\).

\subsection{Proportion of Matrices Meeting Applicability Criteria} \label{ssec:share}

\begin{figure}[tbp]
    \centering
    \includegraphics[width=\linewidth]{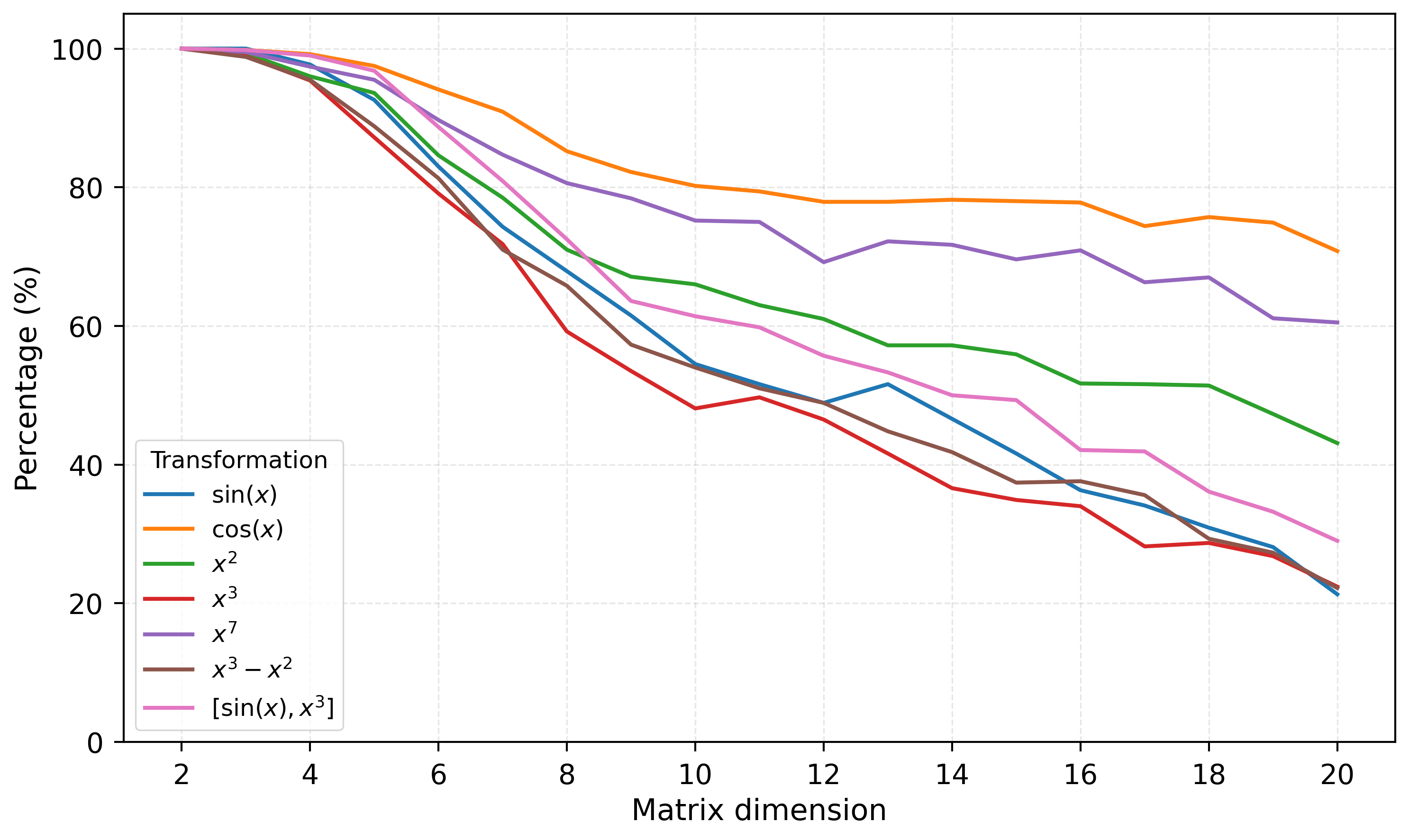}
    \caption{Proportion of experiments where the algorithm is applicable for different transformations and matrix dimensions. Applicability here is decided by whether Neumann series can be used to invert $\bm \Gamma_\rho$ and $\bm R_\pi$ within an experiment.}
    \label{fig:app_share}
\end{figure}

Given the assumptions of our method, it is natural to wonder how frequently precision matrices satisfy its conditions. To evaluate this, we examine the proportion of precision matrices, generated via the Erdős–Rényi model, that satisfy the conditions necessary for employing the Neumann series to approximate the inverses of  $\bm \Gamma_\rho$ and $\bm R_\pi$. This assessment is conducted empirically by sampling and transforming data using the experimental process outlined in Section~\ref{ssec:exp}---specifically Steps~\ref{step:1}(a-f),~\ref{step:2}, and~\ref{step:3}. Figure~\ref{fig:app_share} presents the proportion of experiments in which the algorithm is applicable (i.e., the conditions in Steps~\ref{step:1g} and~\ref{step:4a} are met), as a function of matrix dimension. As seen in the Figure~\ref{fig:app_share}, the proportion decreases as the matrix dimension increases. This trend arises because larger matrices, on average, exhibit higher operator norms, reducing the likelihood that their inverses can be effectively approximated with the Neumann series.

Since different transformations of the data sampled from the same $\bm \Sigma_\rho$ will result in varying $\bm R_\pi$, the rate at which these transformations satisfy the condition also varies. For 10-dimensional data, the algorithm is applicable for approximately 50\%-80\% of cases, depending on the transformation applied. In contrast, for 20-dimensional data, applicability declines to 20\%-70\%. Although all experiments of different transformations approximate the inverse of $\bm \Gamma_\rho$ at a similar rate (since they correspond to Gaussian data), the behavior diverges after transformation to GNPN data. The Neumann series inverts $\bm R_\pi$ more effectively for high-order polynomials and even transformations than for odd transformations. This divergence, evident in Figure~\ref{fig:app_share}, is attributed to the tendency of these transformations to drive correlations in $\bm R_\pi$ closer to zero, thereby increasing the likelihood of successful inversion via the Neumann series.

While applicability of the algorithm diminishes as the dimensionality increases, when the conditions hold, it efficiently (as shown in the next section) recovers the conditional independence structure for a broad class of non-Gaussian distributions. The algorithm's advantage lies in its simplicity: it leverages the precision matrix to infer conditional independence, bypassing explicit density estimation or complex transformations.

\subsection{Sample Efficiency} \label{ssec:efficiency}

\begin{figure}[tbp]
    \centering
    \begin{subfigure}{0.49\textwidth}
        \centering
        \includegraphics[width=\linewidth]{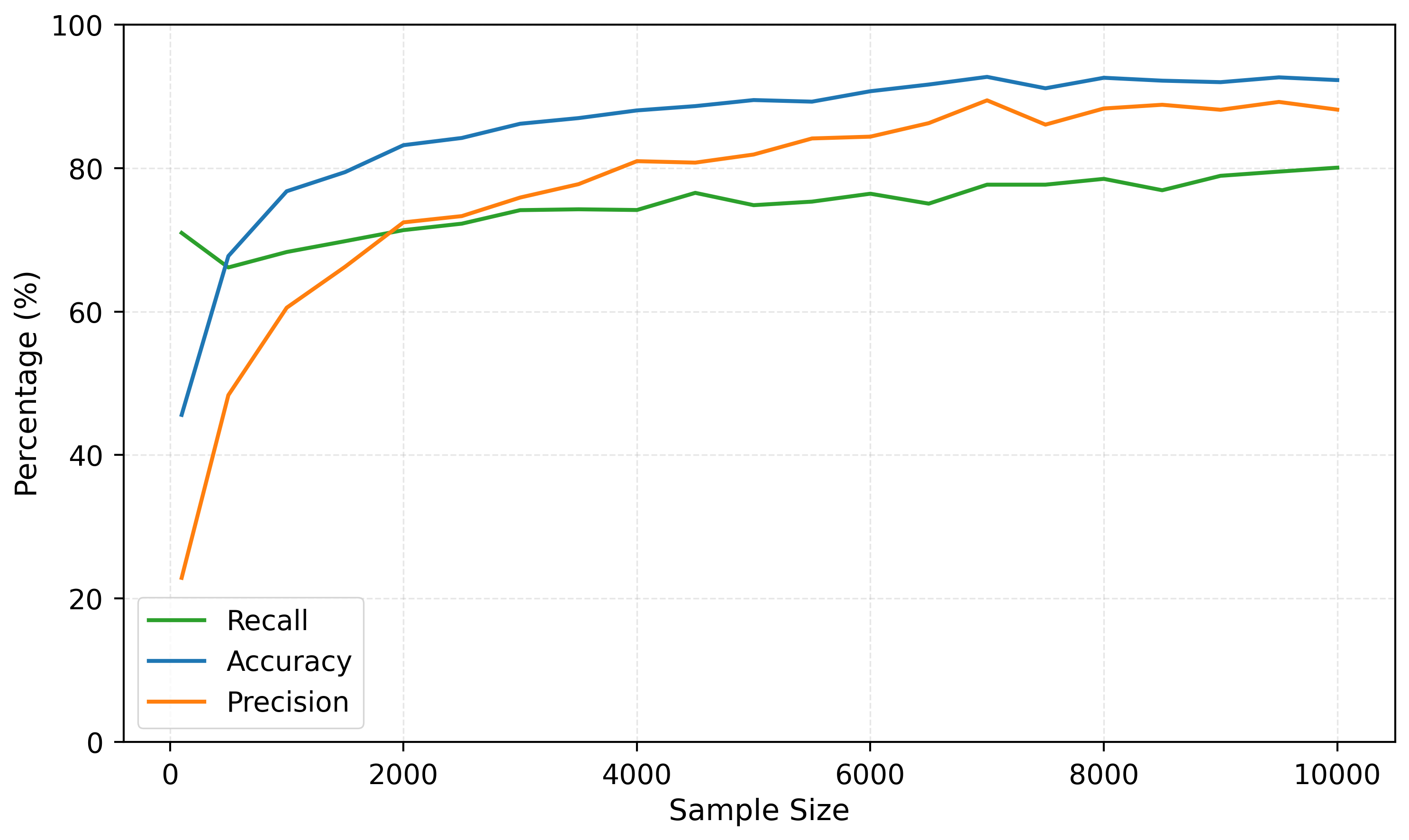}
        \caption{\(x^2\) transformation}
        \label{fig:perf_v_sample_2}
    \end{subfigure}
    \hfill
    \begin{subfigure}{0.49\textwidth}
        \centering
        \includegraphics[width=\linewidth]{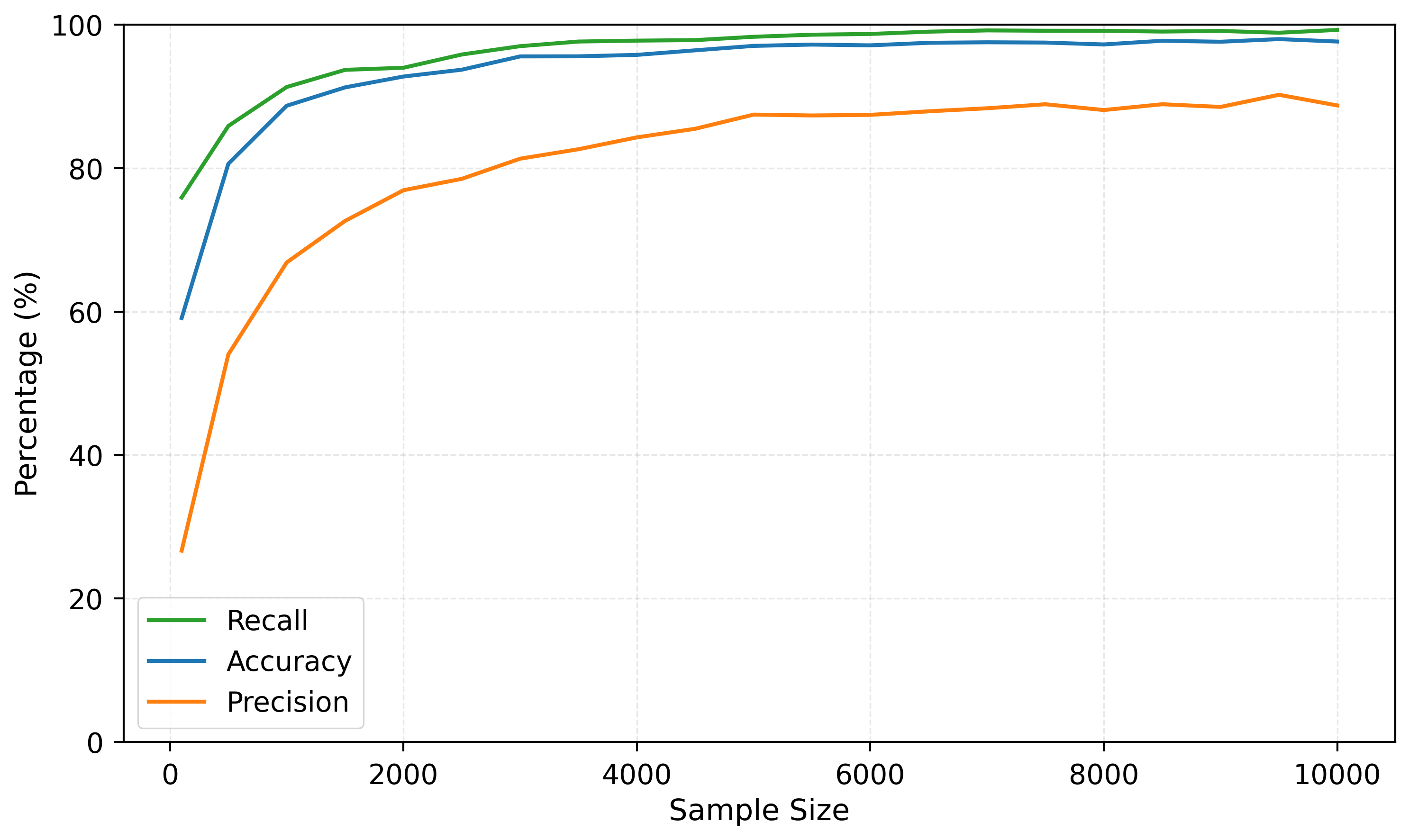}
        \caption{\(x^3\) transformation}
        \label{fig:perf_v_sample_3}
    \end{subfigure}
    \caption{Performance metrics for the conditional independence structure recovered from GNPN data improve rapidly as sample size increases. }
    \label{fig:perf_v_sample}
\end{figure}

Finally, we examine the sample efficiency of the proposed algorithm with sample size $n = 100,\ 500,\ 1000,\ 1500,...10000$ by generating $500$ random matrices for each $n$ and following the same experimentation process described previously. As an example, in Figure~\ref{fig:perf_v_sample}, we show the accuracy, recall, and precision metrics at different sample sizes for an odd and an even transformation. For 10-variate Gaussian data transformed with the \(x^2\) function, the algorithm achieves more than 90\% accuracy and 85\% precision at 6000 samples, and achieves more than 80\% recall at 10000 samples. For the \(x^3\) case, the algorithm achieves 95\% accuracy and recall at 3000 samples, and 85\% precision at 5000 samples. We observe a similar trend in the performance metrics of other transformations, where there is an initial rapid improvement that gradually tapers off as the sample size increases, ultimately converging towards the performance levels shown in Table~\ref{table:1}.

\section{Discussion} \label{sec:dis}
In this work, we present a framework for learning the conditional independence structure of a broad class of non-Gaussian distributions, specifically those derived from diagonal transformations of a Gaussian distribution. Our approach extends previous work by relaxing assumptions about the nature of the transformations, allowing for odd or even functions, a combination of both, or neither. The theoretical contributions of this paper center on the derivation of estimates for the precision matrices of these generalized nonparanormal (GNPN) distributions. The resulting precision matrix retains much of the structure of the original Gaussian, which enables the inference of independence properties directly from this matrix. Specifically, values that were large in magnitude in the Gaussian precision matrix $\bm\Gamma_\rho$ remain comparatively large in the precision matrix of the GNPN data $\bm \Gamma_\pi$, while zero elements in $\bm\Gamma_\rho$ are still small in $\bm\Gamma_\pi$. 

The algorithm we proposed leverages this theory, offering a simple and practical method for recovering the conditional independence structure from generalized nonparanormal data without requiring explicit analysis of the transformation functions. Our experiments demonstrate that the method performs well across a variety of transformations. However, performance suffers for high-degree polynomial transformations as they compress edge weights in $\bm \Gamma_\pi$ closer to zero, making it difficult for the algorithm to differentiate them from noise. 


\acks{We thank Ricardo Baptista for helpful comments on this work. The research reported here was supported in part by the National Science Foundation (NSF grant 2422136).}

\vskip 0.2in
\bibliography{references}

\end{document}